\documentclass[12pt]{article}

\usepackage{scicite}
\usepackage{graphicx}
\usepackage{times}
\usepackage{multirow}
\usepackage{amsmath}
\usepackage{bm}
\usepackage{subcaption}
\usepackage[table,xcdraw,dvipsnames]{xcolor}
\usepackage{float}
\usepackage{placeins}
\usepackage[normalem]{ulem}
\usepackage{newfloat}
\usepackage{hyperref}

\newenvironment{sciabstract}{%
\begin{quote} \bf}
{\end{quote}}

\title{Emergence of Exploratory Look-Around Behaviors through Active Observation Completion~\footnote{This manuscript has been accepted for publication in Science Robotics.  This version has not undergone final editing. Please refer to the complete version of record at \url{https://robotics.sciencemag.org/content/4/30/eaaw6326}. The manuscript may not be reproduced or used in any manner that does not fall within the fair use provisions of the Copyright Act without the prior, written permission of AAAS.}}

\author
{Santhosh K. Ramakrishnan,$^{1,3^{\tiny\dagger\tiny\ddagger}}$, Dinesh Jayaraman,$^{2^{\tiny\dagger}}$, Kristen Grauman$^{1,3}$\\
\\
\normalsize{$^{1}$Department of Computer Science, University of Texas at Austin, Texas}\\
\normalsize{$^{2}$Department of Electrical Engineering and Computer Science, University of California, Berkeley}\\
\normalsize{$^{3}$Facebook AI Research, Austin, Texas}\\
\\
\normalsize{$^\ddagger$To whom correspondence should be addressed; E-mail:  srama@cs.utexas.edu}
\\
\normalsize{$^\dagger$Equal contribution}
}

\date{}

\DeclareMathOperator*{\argmax}{arg\,max}
\newcolumntype{P}[1]{>{\centering\arraybackslash}p{#1}}

\begin{document} 


\baselineskip24pt


\maketitle 

\vspace*{-0.35in}

\begin{sciabstract}
Standard computer vision systems assume access to intelligently captured inputs (e.g., photos from a human photographer),
yet autonomously capturing good observations is a major challenge in itself. We address the problem of learning to
look around: how can an agent learn
to acquire informative visual observations? We propose a reinforcement learning
solution, where the agent is rewarded for reducing its uncertainty about the unobserved portions of its
environment.  Specifically, the agent is trained to select a short sequence of glimpses after which it must infer the appearance of its full environment.  To address the challenge of sparse
rewards, we further introduce
sidekick policy learning, which exploits the asymmetry in observability between training and test time.
The proposed methods learn observation policies that not only perform the  completion task for which they are trained, but also generalize to exhibit useful ``look-around" behavior for a range of active perception tasks.
\end{sciabstract}

\section*{Introduction}
\vspace*{-0.1in}

Visual recognition has witnessed dramatic successes in recent years.  Fueled by benchmarks composed of carefully curated Web photos and videos, the focus has been on inferring semantic labels from human-captured images---whether classifying scenes, detecting objects, or recognizing activities~\cite{ILSVRC15,lin2014microsoft,soomro2012ucf101}. 
However, visual perception requires not only making inferences from observations, but also making decisions about what to observe. Methods that use human-captured images implicitly assume properties in their inputs, such as canonical poses of objects, no motion blur, or ideal lighting conditions. As a result, they gloss over important hurdles for robotic agents acting in the real world.

For an agent, individual views of an environment offer only a small fraction of all relevant information. For instance, an agent with a view of a television screen in front of it may not know if it is in a living room or a bedroom. An agent observing a mug from the side may have to move to see it from above to know what is inside.  

\begin{figure}[ht]
  \centering
  \includegraphics[width=0.7\textwidth]{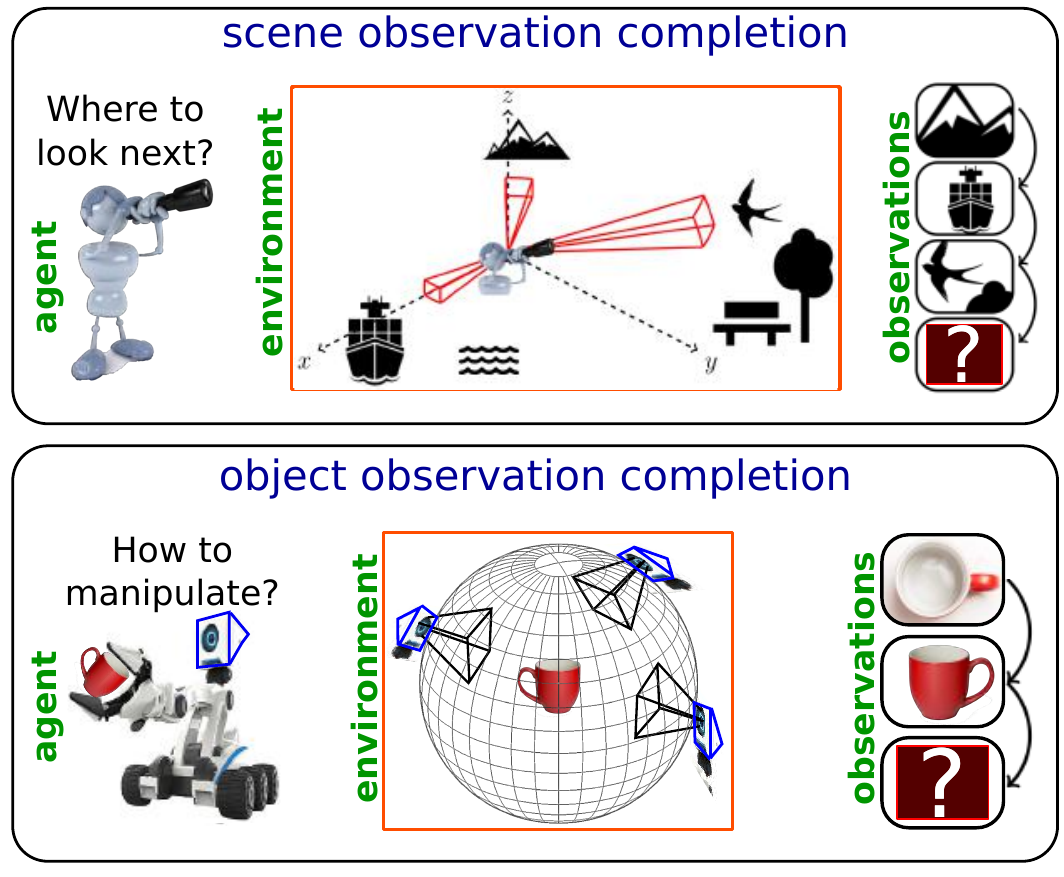}
  \caption{Looking around efficiently is a complex task requiring the ability to reason about regularities in the visual world using cues like context and geometry. Top: An agent that has observed limited portions of its environment can reasonably predict some unobserved portions (e.g., water near the ship), but is much more uncertain about other portions. Where should it look next? Bottom: An agent inspecting a 3D object. Having seen a top view and a side view, how must it rotate the mug now to get maximum new information? Critically, we aim to learn policies that are not specific to a given object or scene, nor to a specific perception task. Instead, the look-around policies ought to benefit the agent exploring new, unseen environments and performing tasks unspecified when learning the look-around behavior.
  }
  \label{fig:conceptfig}
 \vspace{-0.05in}
\end{figure}

An agent ought to be able to enter a new environment or pick up a new object and intelligently (non-exhaustively) ``look around".    
The ability to actively explore would be valuable in both task-driven scenarios (e.g., a drone searches for signs of a particular activity) as well as scenarios where the task itself unfolds simultaneously with the agent's exploratory actions (e.g., a search-and-rescue robot enters a burning building and dynamically decides its mission). 
For example, consider a service robot that is moving around in an open environment without specific goals waiting for future tasks like delivering a package from one person to another or picking up coffee from the kitchen. It needs to efficiently gather information constantly so that it is well-prepared to perform future tasks with minimal delays.
Similarly, consider a search and rescue scenario, where a robot is deployed in a hostile environment such as a burning building or earthquake collapse where time is of the essence. The robot has to adapt to such new unseen environments and rapidly gather information that other robots and humans can use to effectively respond to tasks that dynamically unfold over time (humans caught under debris, locations of fires, presence of hazardous materials). Having a robot that knows how to explore intelligently can be critical in such scenarios as it reduces risks for people while providing an effective response.

Any such scenario brings forth the question of how to collect visual information to benefit  perception. A na\"ive strategy would be to gain full information 
by making every possible observation---that is, looking around in all directions, or systematically examining all sides of an object.  However, observing all aspects is often inconvenient if not intractable. Fortunately, in practice not all views are equally informative.  The natural visual world contains regularities, suggesting not every view needs to be sampled for accurate perception. For instance, humans rarely need to fully observe an object to understand its 3D shape~\cite{soska2008development,soska2010systems,kellman1983perception}, and one can often understand the primary contents of a room without literally scanning it~\cite{torralba2006contextual}. In short, given a set of past observations, some new views are more informative than others. See Figure~1. 

This leads us to investigate the question of how to effectively look around: how can a learning system make intelligent decisions about how to acquire new exploratory visual observations? We propose a solution based on ``active observation completion": an agent must actively observe a small fraction of its environment so that it can predict the pixelwise appearances of unseen portions of the environment. 

Our problem setting relates to but is distinct from prior work in active perception, intrinsic motivation, and view synthesis.  While there is interesting recent headway in active object recognition~\cite{dinesh-eccv2016,germs-bmvc2015,shapenet,ammirato-icra2017} and intelligent search mechanisms for detection~\cite{yeung2016end,mathe-cvpr2016,timely-nips2012}, such systems are supervised and task-specific---limited to accelerating a pre-defined recognition task. 
In reinforcement learning, intrinsic motivation methods define generic rewards such as novelty or coverage~\cite{pathakICMl17curiosity,chen2018learning,Hepp_2018_ECCV}, that encourage exploration for navigation agents, 
but they do not self-supervise policy learning in an observed visual environment, nor do they examine transfer beyond navigation tasks.
View synthesis approaches use limited views of the environment along with geometric properties to generate unseen views~\cite{song2018im2pano3d,ji-cvpr2017,kulkarni-nips2015,jayaraman-eccv2018,gqn-science2018}. Whereas these methods assume individual human-captured images, our problem requires actively selecting the input views themselves.
Our primary goal is not to synthesize unseen views, but rather to use novel view inference as a means to elicit  intelligent exploration policies that transfer well to other tasks.

In the following, we first formally define the learning task,  overview our approach, and present results.  Then after the results, we discuss limitations of the current approach and key future directions, followed by the materials and methods---an overview of the specific deep networks and policy learning approaches we develop.  This article expands upon our two previous conference papers~\cite{dinesh-cvpr2018, santhosh-eccv2018}.

\subsection*{Active observation completion}

Our goal is to learn a policy for controlling an agent's camera motions such that it can explore novel environments and objects efficiently. 
To this end, we formulate an unsupervised learning objective based on active observation completion.
The main idea is to favor sequences of camera motions that will make the unseen parts of the agent's surroundings easier to predict.  The output is a look-around policy equipped to gather new images in new environments.  As we will demonstrate in results, it prepares the agent to perform intelligent exploration for a wide range of perception tasks, such as recognition, light source localization, and pose estimation. 

\subsubsection*{Problem formulation}

The problem setting is formally stated as follows.  The agent starts by looking at a novel environment (or object)
 $X$ from some unknown viewpoint~\cite{footnote1}. 
 It has a budget $T$ of time to explore the environment.  The learning objective is to minimize the error 
 in the agent's  pixelwise reconstruction of the full---mostly unobserved---environment using only the sequence of views selected within that budget. In order to do this, the agent must maintain an internal representation of how the environment would look  conditioned on the views it has seen so far.
 
We represent the entire environment as a ``viewgrid" containing views from a discrete set of viewpoints. To do this, we evenly sample $N$ elevations from $-90^{\circ}$ to $90^{\circ}$ and
$M$ azimuths from $0^{\circ}$ to $360^{\circ}$ and form all $MN$ possible (elevation, azimuth) pairings. The viewgrid is then denoted by $V(X) = \{x(X, \theta^{(i)}) | 1 \le i \le MN \}$, 
where $x(X, \theta^{(i)})$ is the 2D view of $X$ from viewpoint $\theta^{(i)}$ which is the $i^{th}$ pairing.   More generally, $\theta^{(i)}$ could capture 
both camera angles and position; however, to best exploit existing datasets, we limit our experiments to camera rotations alone with no translation movements. 

The agent expends its time budget $T$ in discrete increments by selecting $T-1$ camera motions in sequence.  Each camera motion comprises an actively chosen ``glimpse". 
At each time step, the agent gets an image observation $x_{t}$ from the current viewpoint. It then makes an exploratory motion ($a_{t}$) based on its policy $\pi$. When the agent executes
action $a_{t} \in \mathcal{A}$, the viewpoint changes according to $\theta_{t+1} = \theta_{t} + a_{t}$. For each camera motion $a_{t}$
executed by the agent, a reward $r_{t}$ is provided by the environment.  
Using the view $x_{t}$, the agent updates its internal representation
of the environment, denoted $\hat{V}(X)$.  Because camera motions are restricted to have proximity to the current camera angle 
and candidate viewpoints partially overlap, the discrete action space promotes efficiency without neglecting the physical realities of the problem (following~\cite{germs-bmvc2015,dinesh-eccv2016,dinesh-cvpr2018,johns-cvpr2016}). During training, the full viewgrids of the 
environments are available to the agent as supervision. During testing, the system must predict the complete viewgrid, having seen only a few views within it.

We explore our idea in two settings. See Figure 1. In the first, the agent scans a scene through its limited field-of-view camera; the goal is to select efficient camera motions so that after a few glimpses, it can model unobserved portions of the scene well. In the second, the agent manipulates a 3D object to inspect it; the goal is to select efficient manipulations so that after only a small number of actions, it has a full model of the object's 3D shape. In both cases, the system must learn to leverage visual regularities (shape primitives, context, etc.) that suggest the likely contents of unseen views, focusing on portions that are hard to ``hallucinate" (i.e., predict pixelwise).

Posing the active view acquisition problem in terms of observation completion has two key advantages:  generality and low cost (label-free) training data.
The objective is general, in the sense that pixelwise reconstruction places no assumptions about the future task for which the glimpses will be used. 
The training data is low-cost, since no manual annotations are required;
the agent learns its look-around policy by exploring any visual scene or object.  This assumes that capturing images is much more cost-effective than manually annotating images.

\begin{figure}[ht]
    \centering
    \includegraphics[width=0.9\textwidth, trim={0 0 0 0}, clip]{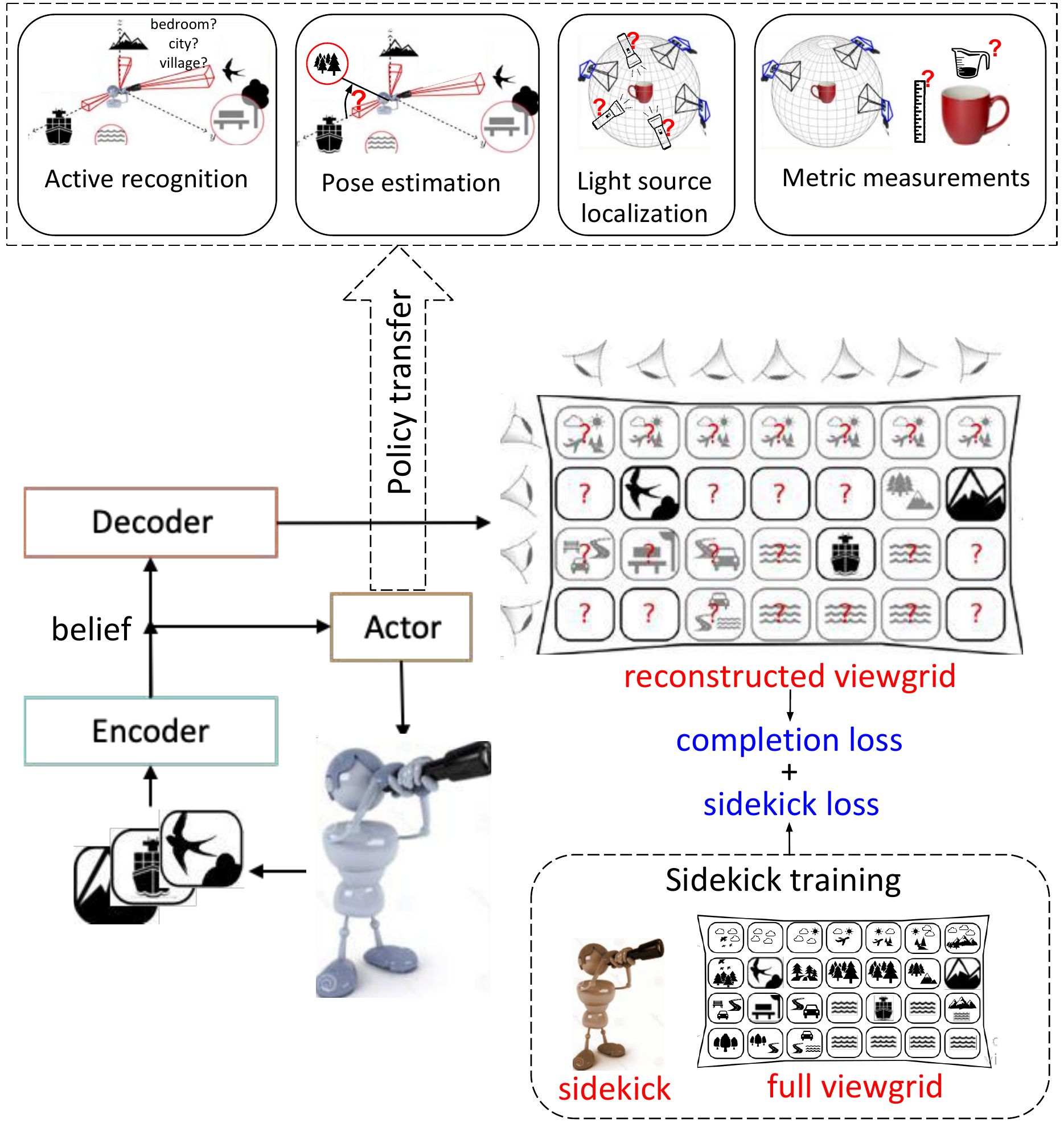}
    \caption{Approach overview: The agent (actor) encodes individual views from the environment and aggregates them into a belief state vector. This belief is used by the decoder to get the reconstructed viewgrid. The agent's incomplete belief about the environment leads to uncertainty over some viewpoints (red question marks). To reduce this uncertainty, the agent intelligently samples more views based on its current belief within a fixed time budget $T$. The agent is penalized based on the reconstruction error at the end of $T$ steps (completion loss). Additionally, we provide guidance through sidekicks (sidekick loss) which exploit the full viewgrid---only at training time---to alleviate uncertainty in training due to partial observability. The learned exploratory policy is then transferred to other tasks (top row shows four tasks we consider).}
    \label{fig:approach}
\end{figure}

\subsection*{Approach overview}\vspace{-0.2cm}

The active observation completion task poses three major challenges.  Firstly, to predict unobserved views well, the agent must learn to understand 3D relationships from very few views.  Classic geometric 
solutions struggle under these conditions. Instead, our reconstruction must draw on semantic and contextual cues. Secondly, intelligent action selection is critical to this task.  Given a set of past 
observations, the system must act based on {which} new views are likely to be most informative, i.e., determine which views would most improve its model of the full viewgrid. We stress that
the system will be faced with objects and scenes it has never encountered during training, yet still must intelligently choose where it would be valuable to look next. 

\vspace{-0.2cm} As a core solution to these challenges, we present a reinforcement learning (RL) approach for
active observation completion~\cite{dinesh-cvpr2018}. See Figure 2.  Our RL approach uses a recurrent neural network to aggregate information over a sequence of views; a stochastic neural network uses that aggregated state and current observation to select a sequence of useful camera motions.  The agent is rewarded based on its predictions of unobserved views.  It therefore learns a policy to intelligently select actions (camera motions) to maximize the quality of its predictions.  During training, the complete viewgrid is known, thereby allowing the agent to ``self-supervise" its policy learning, meaning it learns without any human-provided labels.  See Materials and Methods below for the details of our approach.

We judge the quality of viewgrid reconstruction in the pixel space so as to maintain generality: all pixels for the full scene (or 3D object) would encompass all potentially useful visual information for any task. 
Hence, our approach avoids committing to any intermediate semantic representation, in favor of learning policies that seek 
generic information useful to many tasks. That said, our formulation is easily adaptable to more specialized settings.  For example, if the target tasks only require semantic segmentation labels, 
the predictions could be in the space of object labels instead.\vspace{-0.1cm}

Reinforcement learning approaches often suffer from costly exploration stages and partial state observability. In particular, an active visual agent has to take a long series of actions purely based on the limited information available from its first person view~\cite{zhu-iccv2017,gupta2017unifying,zhu2017target,dinesh-cvpr2018}.
The most effective viewpoint trajectories are buried among many mediocre ones, impeding the agent's exploration in complex state-action spaces.\vspace{-0.1cm}

To address this challenge, as the second main technical contribution of this work, we introduce ``sidekick 
policy learning".  In the active observation completion task there is a natural asymmetry in observability: once deployed an active exploration agent can only move the camera to look around nearby, yet during training it can access omnidirectional viewpoints.
Existing methods facing this asymmetry simply restrict the agent to the same partial observability during training~\cite{shapenet,johns-cvpr2016,dinesh-eccv2016,dinesh-cvpr2018,zhu-iccv2017,dinesh-pami2018}.
In contrast, our sidekick approach introduces reward shaping and demonstrations that leverage full observability during training to precompute the information content of each candidate glimpse. The sidekicks then guide the agent to visit information hotspots in the environment or sample information-rich trajectories, while accounting for the fact that observability is only partial during testing~\cite{santhosh-eccv2018}.
By doing so, sidekicks accelerate the training of the actual agent and improve the overall performance.
We use the name ``sidekick" to signify how a sidekick to a hero (e.g., in a comic or movie) provides alternate points of view, knowledge, and skills that the hero does not have.  In contrast to an ``expert"~\cite{guo2014deep,vapnik2016learning}, 
a sidekick complements the hero (agent), yet 
cannot solve the main task at hand by itself.   See Materials and Methods below for more details. 

We show that the active observation completion policies learned by our approach serve as exploratory policies that are  transferable to entirely new  tasks and environments. Given a new task, rather than train a policy with task-specific rewards to direct the camera, we drop in the pre-trained look-around policy.  We demonstrate that policies learned via active observation completion transfer well to several semantic and geometric estimation tasks, and they even perform competitively with supervised task-specific policies (please see the look-around policy transfer section in Results).

\FloatBarrier
\section*{Results}

We next present experiments to evaluate the behaviors learned by the proposed look-around agents.

\subsection*{Datasets}
\label{sec:dataset}
For benchmarking and reproducibility, we evaluate our approach on two widely used datasets:

\subsubsection*{SUN360 Dataset for Scenes\vspace{-1.0em}}
For this dataset, our limited field-of-view ($60^{\circ}$) agent attempts to complete an omnidirectional scene. SUN360~\cite{xiao2012recognizing} has spherical panoramas of 26 diverse categories. The dataset consists of 6,174 training, 1,013 validation and 1,805 testing examples.
The viewgrid has 32$\times$32 resolution 2D images sampled from $M=4$ camera elevations ($-67.5^{\circ},-22.5^{\circ},$ $22.5^{\circ},67.5^{\circ}$) and $N=8$ azimuths ($45^{\circ},90^{\circ},\ldots,360^{\circ}$). 

\subsubsection*{ModelNet Dataset for Objects\vspace{-1.0em}} 
For this dataset, our agent manipulates a 3D object to complete its viewgrid of the object seen from all viewing directions.  The viewgrid constitutes an implicit image-based 3D shape model.  ModelNet~\cite{shapenet} has two subsets of CAD models: ModelNet-40 (40 categories) and ModelNet-10 (a 10 category-subset of ModelNet-40). Excluding the ModelNet-10 classes, ModelNet-40 consists of 6,085 training, 327 validation and 1,310 testing examples. ModelNet-10 consists of 3,991 training, 181 validation and 727 testing examples.
The viewgrid has 32$\times$32 resolution 2D images sampled from $M=6$ camera elevations ($-75^{\circ},-45^{\circ},\ldots,45^{\circ},75^{\circ}$) and $N=10$ azimuths ($20^{\circ},56^{\circ},92^{\circ},\ldots,344^{\circ}$)~\cite{footnote2}. 
We render the objects using substantial lighting variations in order to increase  difficulty in perception.  To test the agent's ability to generalize to previously unseen categories, we always test on object categories in ModelNet-10, which are unseen during training.

For both datasets, at each timestep the agent moves within a 5 elevations$\times$5 azimuths neighborhood from the current position. Requiring nearby motions reflects that the agent cannot teleport, and it ensures that the actions have approximately uniform real-world cost.  Balancing task difficulty (harder tasks require more views) and training speed (fewer views is faster) considerations, we set the training episode length $T=4$ a priori. 
By training for a target budget $T$, the agent learns non-myopic behavior to best utilize the expected exploration time.  Note that while increasing $T$ during training increases training costs considerably, doing so
can naturally lead to better reconstructions (please see Supplementary for longer episode results).

\subsection*{Baselines}
We test our active completion approach with and without sidekick policy learning~\cite{footnote3} 
---\texttt{\small lookaround} and \texttt{\small lookaround+spl}, respectively---compared to a variety of baselines:
\begin{itemize}
\itemsep-0.5em
  \item \texttt{\small one-view} is our method trained with $T=1$. No information aggregation or action selection is performed by this baseline.
  \item \texttt{\small rnd-actions} is identical to our approach, except that the action selection module is replaced by randomly selected actions from the pool of all possible actions.
  \item \texttt{\small large-action} chooses the largest allowable action repeatedly. This tests if far-apart views are sufficiently informative.
  \item \texttt{\small peek-saliency} moves to the most salient view within reach at each timestep, using a popular saliency metric~\cite{harel2006graph}. To avoid getting stuck in a local saliency maximum, it does not revisit seen views. Note that this baseline peeks at neighboring views prior to action selection to measure saliency, giving it an unfair and impossible advantage over our methods and the other baselines.
\end{itemize}
These baselines all use the same network architecture as our methods, differing only in the exploration policy which we seek to evaluate. In the interest of evaluating on a wide range of starting positions, we evaluate each method $MN$ times on each test viewgrid, starting from all possible view points.

\subsection*{Active observation completion results}

\begin{figure}
\begin{subfigure}{\textwidth}
    \centering
    \includegraphics[width=\textwidth, trim={0 0 0 0}, clip]{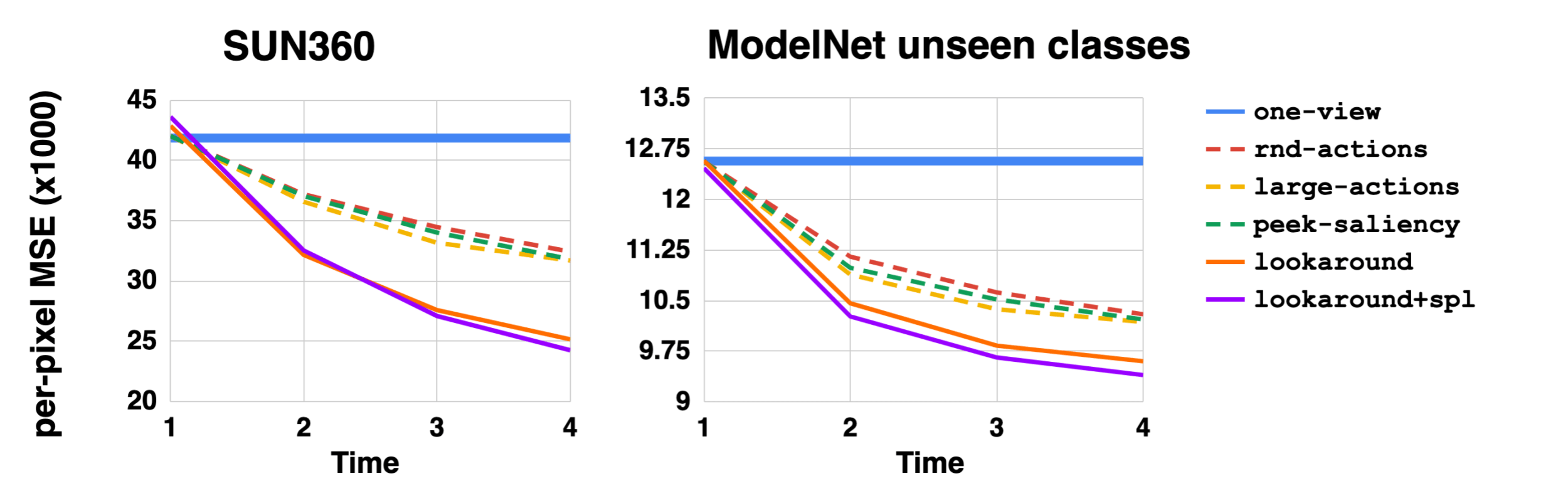}
    \caption{\footnotesize Pixelwise MSE errors vs. time on both datasets.}\vspace*{0.15in}
    \label{fig:look_around_quant}
\end{subfigure}

\begin{subfigure}{\textwidth}
\centering
\resizebox{\textwidth}{!}{%
\begin{tabular}{| p{4cm} || P{3em} P{3.5em} | P{3em} P{3em}|| P{3em} P{3em} | P{3em} P{3em} |}
\hline
\multirow{2}{*}{Method}                    & \multicolumn{4}{c||}{SUN360}                                         & \multicolumn{4}{c|}{ModelNet unseen classes}                                       \\ \cline{2-9} 
                                           & \multicolumn{2}{c|}{average}     & \multicolumn{2}{c||}{adversarial} & \multicolumn{2}{c|}{average}     & \multicolumn{2}{c|}{adversarial} \\
                                           & mean $\downarrow$ & $\%\uparrow$ & mean $\downarrow$ & $\%\uparrow$ & mean $\downarrow$ & $\%\uparrow$ & mean $\downarrow$ & $\%\uparrow$ \\ \hline
\texttt{\small one-view}                                &      41.85        &      -       &        70.44      &       -      &       12.57       &      -       &       20.32         &      -                   \\
\texttt{\small rnd-actions}                             &      32.42        &     22.52    &        54.66      &     22.39    &       10.30       &     18.06    &       14.01         &     31.04             \\
\texttt{\small large-actions}                           &      31.68        &     24.28    &        42.56      &     39.57    &       10.18       &     18.97    &       13.04         &     35.79             \\
\texttt{\small peek-saliency}                           &      31.80        &     24.02    &        45.24      &     35.76    &       10.22       &     18.67    &       12.78         &     37.05             \\
\texttt{\small lookaround}       &      25.14        &     39.91    &  \textbf{30.44}   &\textbf{56.79}&        9.60       &     23.60    &       12.01         &     40.88             \\
\texttt{\small lookaround+spl} &   \textbf{24.24}  &\textbf{42.06}&        30.75      &     56.34    &   \textbf{9.40}   &\textbf{25.23}& \textbf{11.85}      & \textbf{41.67}    \\
\hline
\end{tabular}
}
\caption{Average/adversarial MSE error $\times 1000$ ($\downarrow$ lower is better) and corresponding improvements (\%) over the \texttt{one-view} model ($\uparrow$ higher is better) on both datasets.}
\label{tab:quant_active_obs}
\end{subfigure}
\caption{Scene and object completion accuracy under different agent behaviors.  Top plots (a) show error rates over time as more glimpses are acquired, and bottom table (b) shows errors/improvements after all $T$ glimpses are acquired.}
\end{figure}

We show the results of scene and object completion on SUN360 and ModelNet (unseen classes) in Figure 3b. The metrics ``average" and ``adversarial" measure the expected value of the average and maximum pixelwise mean squared errors (MSE) over all starting points for a single sample, respectively. While the former measures the average expected performance, the latter measures the worst-case performance when starting from the hardest place in each sample (averaged over examples). 
We additionally report the relative improvement of each model over \texttt{\small one-view} in order to isolate the gains obtained due to action selection over a pre-trained $T=1$ model. 
Since all methods share the same pre-training stage of \texttt{\small one-view}, this metric provides an apples-to-apples measure of how well the different strategies for moving perform.
All methods are evaluated over $T=4$ time steps in accordance with the training budget unless stated otherwise. 

As expected, all methods that acquire multiple glimpses 
outperform \texttt{\small one-view} by taking advantage of the extra information that is available from additional views. Both of our approaches \texttt{\small lookaround} and \texttt{\small lookaround+spl} significantly outperform the others on all settings. The \texttt{\small peek-saliency} agent hovers near the most salient views in the neighborhood of the starting view since nearby views tend to have similar saliency scores. The \texttt{\small large-actions} agent's accuracy often tends to saturate near the top or bottom of the viewgrid after reaching the environment boundaries. Compared to these behaviors, intelligent sampling of actions using our learned policy leads to significant improvements. Using sidekicks in \texttt{\small lookaround+spl} improves performance and convergence speed. This is consistent with our results reported in~\cite{santhosh-eccv2018} and demonstrates the advantage of using sidekicks. The faster convergence of \texttt{\small lookaround+spl} is shown in the Supplementary Materials.

Whereas Figure 3b shows the agents' ultimate ability to infer the entire scene (object), Figure 3a shows the reconstruction errors as a function of time.  As we can see, the error reduces consistently over time for all methods, but it drops most sharply for \texttt{\small lookaround} and \texttt{\small lookaround+spl}. Faster reduction in the reconstruction error indicates more efficient information aggregation. 

Visualizations of the agent's evolving internal belief state echo this quantitative trend.  
Figure 4 shows observation completion episodes from the \texttt{\small lookaround} agent along with the ground truth viewgrid, viewing angles selected by the agent, and reconstruction errors over time.  We show the SUN360 viewgrids in equirectangular projection for better visualization.  
Initially, the agent exhibits significant uncertainty in its belief, as seen in the poorly decoded reconstructions and large MSE values. However, over time, it actively samples views that quickly improve the reconstruction quality. 

Figures 5 and 6 visualize the ultimate reconstructions after all $T$ glimpses are acquired~\cite{footnote4}. 
For contrast, we also display the results for \texttt{\small rnd-actions} in Figure 5. The policies learned by our agent lead to more realistic and accurate reconstructions.
Though the agent only sees about 15\% of all the pixels, its choice of informative glimpses allows it to anticipate the remainder of the novel scene or object.  Movie S1 in the Supplementary Materials shows walkthroughs of the reconstructed environments from the agent's egocentric point of view.

\begin{figure}
    \centering
    \includegraphics[width=\textwidth, trim={0 0 0 0}, clip]{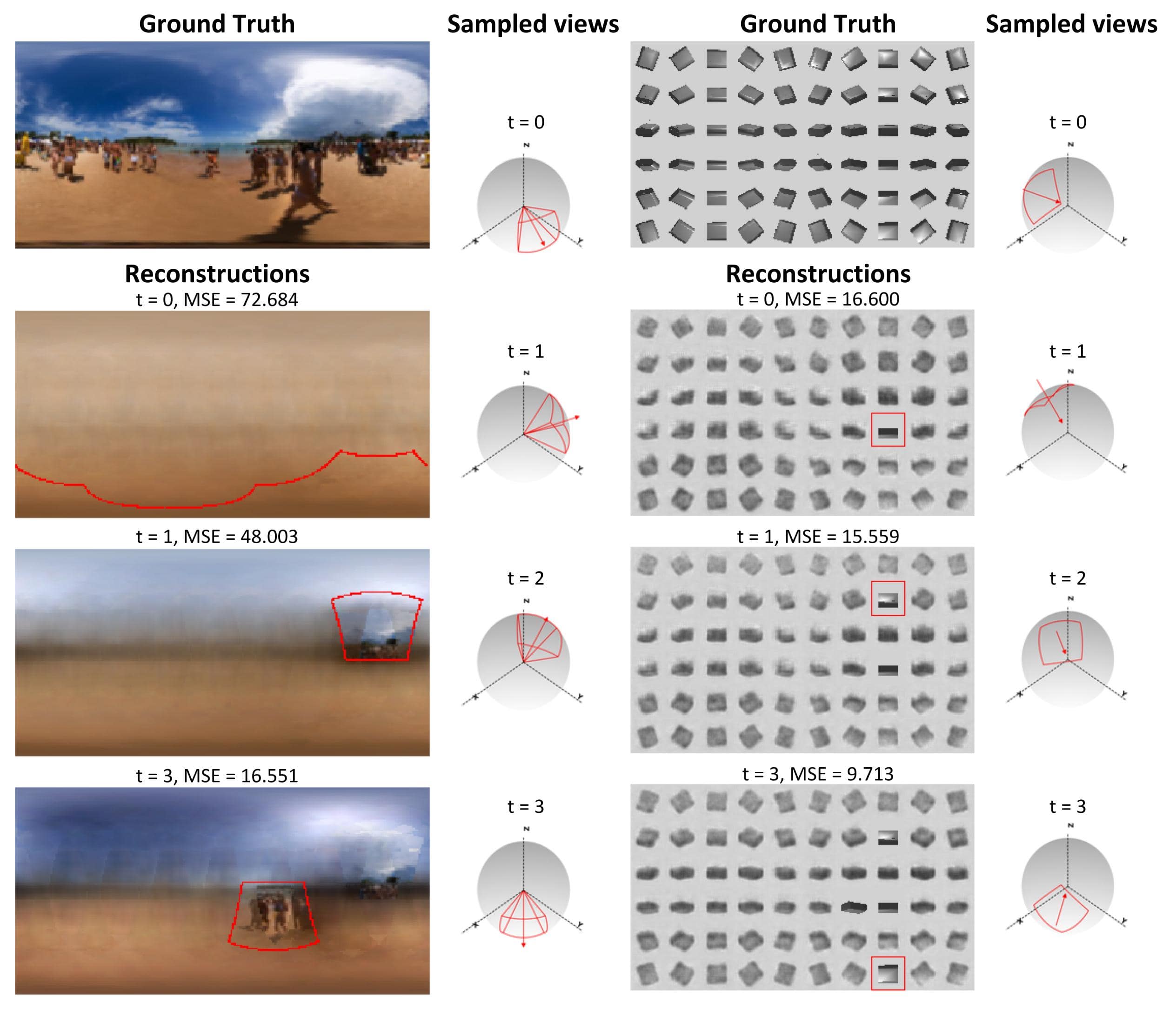}
    \caption{Episodes of active observation completion for SUN360 (left) and ModelNet (right). For each example, the first row on the left shows the ground-truth viewgrid, the subsequent rows on the left show the reconstructions at times $t=0, 1, T-1=3$ along with the pixelwise MSE error ($\times 1000$) and the agent's current glimpse (marked in red). On the right, the sampled viewing angles of the agent at each time step are shown on the viewing sphere (marking the agent's viewpoint and field-of-view using a red arrow and outline on the sphere). The reconstruction quality improves over time as it quickly refines the scene structure and object shape.}
    \label{fig:look_around_qual_time}
\end{figure}

\begin{figure}
    \centering
    \includegraphics[width=\textwidth, trim={0 0 0 0}, clip]{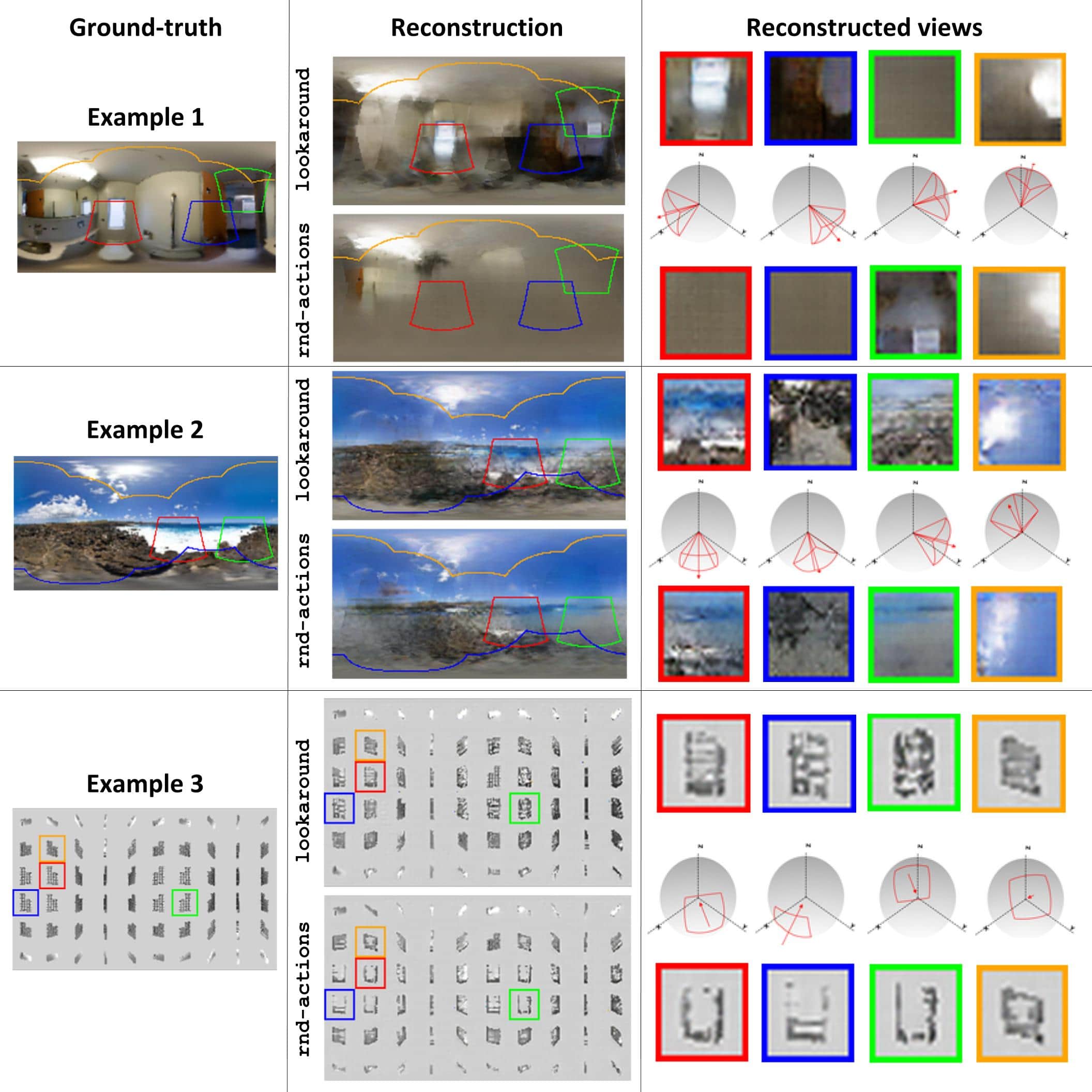}
    \caption{Three examples of reconstructions after $T=6$ glimpses (in order to generate more complete images).
    The first column shows the ground-truth viewgrids (equirectangular projections for SUN), the second column shows the corresponding GAN-refined reconstructions of \texttt{\small lookaround} and \texttt{\small rnd-actions} agents, and the third column shows handpicked unseen views (marked on the ground-truth) and the corresponding angles. Please see Supplementary for more GAN refinement details.  Best viewed on PDF with zoom. Using an intelligent policy, \texttt{\small lookaround} captures more information from the scene leading to more realistic reconstructions (examples 1 and 3). While \texttt{\small rnd-actions} leads to realistic reconstructions on example 2, its textures and content differ from the ground truth, especially on the ground. Note that the bounding boxes over views are warped to curves on the equirectangular projection for SUN360. 
    } 
    \label{fig:look_around_qual_gan}
\end{figure}

\begin{figure}
    \centering
    \includegraphics[width=1.1\textwidth, trim={0 0 0 0}, clip]{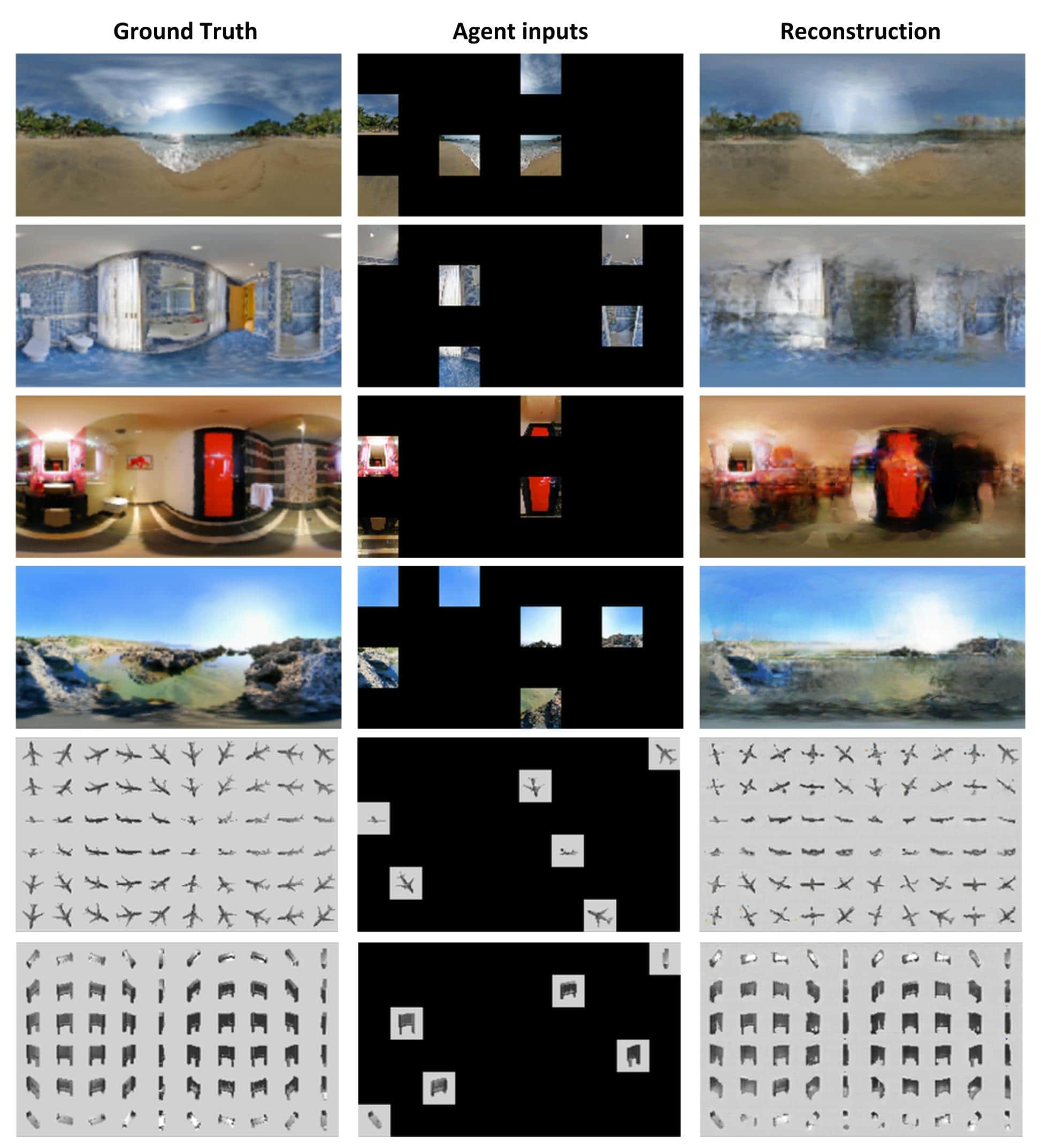}
    \caption{The ground truth 360 panorama or viewgrid, agent glimpse inputs, and final GAN-refined reconstructions for multiple environments from SUN360 and ModelNet.  See also the video provided in the Supplementary Materials.} 
    \label{fig:gan_large_examples}
\end{figure}

\subsection*{Look-around policy transfer}
Having shown that our unsupervised approach successfully trains  policies to acquire visual observations useful for completion, we next test how well the policies transfer to new tasks. 
Recall, our hypothesis is that the glimpses acquired to maximize completion accuracy will transfer well to solve perception tasks efficiently, since they are chosen to reveal maximal information about the full environment or object.

To demonstrate transfer, we first train a \texttt{\small rnd-actions} model for each of the target tasks  (``model A'') and a \texttt{\small lookaround} model for the active observation completion task (``model B''). The policy from model B is then used to select actions for the target task using model A's task head (see details in the unsupervised policy transfer section in Materials and Methods).  In this way, the agent learns to solve the task given arbitrary observations, then inherits our intelligent look-around policy to (potentially) solve the task more quickly---with fewer glimpses.  The successful outcome will be if the  look-around agent can solve the task with similar efficiency as a supervised task-specific policy, despite being unsupervised and task-agnostic.
We test policy transferability for the following four tasks.

\subsubsection*{Task 1: Active categorization}
The first task is category recognition: the agent must produce the category name of the object or scene it is exploring.
We plug look-around policies into the active categorization system from~\cite{dinesh-eccv2016} and follow a similar setup.  
For ModelNet, we train model A on ModelNet-10 training objects, and the active observation completion model (model B) on ModelNet-40 training objects, which are disjoint classes from those in the target ModelNet-10 dataset. For SUN360, both models are trained on SUN360 training data. We replicate the results from~\cite{dinesh-eccv2016} and use the corresponding architecture and training strategies. In particular, the classification head is trained with a cross-entropy loss over the set of classes and the supervised reward function for policy learning is the negative of the classification loss at the end of the episode.  We refer the readers to~\cite{dinesh-eccv2016} for the full details. Performance is measured using classification accuracy on the test set.

\subsubsection*{Task 2: Active surface area estimation}
The second task is surface area estimation. The agent
starts by looking at some view of the object and must intelligently select subsequent viewing angles to estimate the 3D object's surface area. The task is relevant for a robot that needs to interact with an unfamiliar object.
The 3D models from ModelNet-10 are converted into 50x50x50 voxel occupancy grids. 
The true surface area is the number of unoccupied voxels that are adjacent to occupied voxels. Estimation is posed as a regression task where the agent predicts a normalized metric value between 0 and 1. Performance is measured using the relative MSE between predicted and ground truth areas on the test set, i.e, if the ground truth and predicted areas are $m_{\text{g}}$, $m_{\text{p}}$ respectively, the error for one example is $\big((m_{\text{g}}-m_{\text{p}})/m_{\text{g}}\big)^2$. This normalizes the error so that it remains comparable across objects of different sizes.

\subsubsection*{Task 3: Active light source localization}
In the third task, the agent is required to localize the sources of light present surrounding the 3D object. To design a controlled experimental setting, when rendering the ModelNet objects, we  place a single light source randomly at any one of two possible azimuths and four possible elevations relative to the object (see Figure S2 in Supplementary). The task is posed as a four-way classification problem where the agent is required to identify the correct elevation (irrespective of the azimuth, such that there can be no unfair orientation bias). Performance is measured using localization accuracy on the test set.

\subsubsection*{Task 4: Active pose estimation}
The fourth task is camera pose estimation.
Having explored the environment, the agent is required to identify the elevation and relative azimuth of a given reference view. We propose a simple solution to this problem. By using the agent's reconstruction after $T$ time-steps, we measure the $\ell_2$-distance between the given view and each of the reconstructed views. The elevation and azimuth of the reconstructed view leading to the smallest $\ell_2$-distance is predicted as the pose. The agent uses its own decoder as opposed to the decoder from \texttt{\small rnd-actions} as done in previous tasks. We do not evaluate pose estimation on ModelNet due to the ambiguity arising from symmetric objects. The models are evaluated using the absolute angular error (AE) in (1) elevation and (2) azimuth predictions, denoted by `AE azim.' and `AE elev.' in Table 1. During evaluation, the starting positions of the agent are selected uniformly over the grid of views. The reference view is sampled randomly from the viewgrid for each episode.\\

\noindent For baselines, we use  \texttt{\small one-view}, \texttt{\small rnd-actions}, \texttt{\small large-action}, \texttt{\small peek-saliency} (defined in the previous section) and \texttt{\small supervised}. \texttt{\small supervised} is a policy that is trained specifically on the training objective for each task, i.e., with task-specific rewards. 

We compare the transfer of \texttt{\small lookaround} and \texttt{\small lookaround+spl} to these baselines in Table 1. The transfer performance of our policies is better than that of \texttt{\small rnd-actions} on all tasks. This shows that intelligent sequential camera control has scope for improving these perception tasks' efficiency.
Overall, our look-around policy transfers well across tasks,  competing with or even outperforming the supervised task-specific policies.  Furthermore, our look-around policies consistently perform the tasks better than the baseline policies for glimpse selection based on saliency or large actions.

For active recognition on ModelNet, most of the methods perform similarly.  On that dataset, recognition with a single view is already fairly high, leaving limited headroom for improving with additional views, intelligently selected or otherwise.
On pose estimation, our learned policies outperform the baselines as expected since the reconstructions generated by our agents are more accurate. On light source localization, our policies show competitive results and come close to the performance of \texttt{\small supervised}. They also significantly outperform the remaining baselines, demonstrating successful transfer. For surface area estimation, we observe that all methods, including the supervised policies, manage only marginal gains over \texttt{\small one-view}. 
We believe that this is an indication of the difficulty of these tasks, as well as the necessity for more 3D-specific architectures such as those that produce voxel grids, point clouds, or surface meshes as output~\cite{choy20163d,Fan_2017_CVPR,wang2018pixel2mesh}. 

These results demonstrate the effectiveness of learning active look-around policies via observation completion on unlabelled datasets---without task-specific rewards. As we see in Table 1, such policies can successfully transfer to a wide range of perception tasks and often perform on par with supervised task-specific policies.

\begin{table}[t]
\centering
\resizebox{\textwidth}{!}{%
\begin{tabular}{|l||c|c|c||c|c|c|}
\hline
                                            & \multicolumn{3}{c||}{SUN360}                                     & \multicolumn{3}{c|}{ModelNet}                                                                                 \\
\hline             
\multicolumn{1}{|r||}{Task}                 & Active recogn.     & \multicolumn{2}{c||}{Pose estimation.}      & Active recogn.       & Light source loc.    & Surface area                   \\
Method                                      & Accuracy$\uparrow$ & AE azim.$\downarrow$ & AE elev.$\downarrow$ & Accuracy$\uparrow$   & Accuracy $\uparrow$  & RMSE$\times 100 $ $\downarrow$ \\
\hline                                                                                                                                 
\texttt{\small one-view}                           & 51.94              & 75.74                & 30.32                & 83.60                & 58.74                & 21.22                          \\
\texttt{\small rnd-actions}                        & 62.90              & 66.18                & 19.53                & 88.46                & 72.97                & 19.04                          \\
\texttt{\small large-action}                       & 63.73              & 67.57                & 19.94                & 89.05                & 75.14                & 18.38                          \\
\texttt{\small peek-saliency}                      & 64.20              & 65.46                & 19.76                & 88.74                & 71.19                & 18.85                          \\
\texttt{\small supervised}                         & 68.21              & 51.36                &  9.81                & 88.58                & \textbf{86.30}       & 18.43                          \\
\texttt{\small lookaround}                         & 68.89              & 50.00                &  9.94                & 89.00                & 83.29                & 18.82                          \\
\texttt{\small lookaround+spl}                     & \textbf{69.32}     & \textbf{47.13}       & \textbf{9.36}        & \textbf{89.38}       & 83.08                & \textbf{18.14}                          \\
\hline

\end{tabular}%
}
\caption{Transfer results: \texttt{\small lookaround} and \texttt{\small lookaround+spl} are transferred to the \texttt{\small rnd-actions} task-heads from each task.  The same unsupervised look-around policy successfully accelerates a variety of tasks---even competing well with the fully supervised task-specific policy (\texttt{\small supervised}).}
\label{tab:quant_transfer}
\end{table}

\section*{Conclusion}

We propose the task of active observation completion to facilitate learning look-around behaviors in a task-independent way. Our proposed approach outperforms several baselines and effectively anticipates the high-level properties of the environment, having observed only a small fraction of the scene or 3D object.  We further show that adding the proposed RL sidekicks leads to faster training and convergence to better policies (Figures 3 and S3). Once look-around behaviors are learned, we show that they can be effectively transferred to a wide range of semantic and geometric tasks where they at times even outperform supervised policies trained in a traditional task-specific manner (Table 1).

While we are motivated to devise sidekick policy learning for active visual exploration, it is more generally applicable whenever an RL agent can access greater observability during training than during deployment.  
For example, agents may operate on first-person observations during test-time, yet have access to  multiple sensors during training in simulation environments~\cite{dosovitskiy2017carla,pinto2017asymmetric,embodiedqa}.
Similarly, an active object recognition system~\cite{dinesh-eccv2016,johns-cvpr2016,ammirato-icra2017,shapenet,dinesh-pami2018} can only see its previously selected views of the object; yet if trained with CAD models, it could observe all possible views while learning.  Future work can explore sidekicks in such scenarios.

Despite the promising results, our approach does have several shortcomings and our work points to several interesting directions for future work.  
While the agent is moving from one view to another, it does not use the information available during this motion. This is reasonable since allowable actions are confined to a neighborhood of the current observation, and hence relatively close in 3D world space. Still, an interesting setting would be to use the sequence of views obtained while the action is being executed. 

Secondly, our current action space is discretized to promote training efficiency, and we assume that each action has unit cost and optimize the agent to perform well for a fixed cost budget.  The unit cost is approximately correct given the locality of the action space.  Nonetheless, it could be interesting to adapt to
free-range actions with action-specific costs by allowing the agent to sample any action (continuous or discrete) and penalizing it based on the cost of that action.  Such costs could be embodiment-specific. For example, humanoid robots may find it easier to move forward when compared to turning and walking, whereas wheeled robots can perform both motions equally well. Such a formulation would also naturally account for 
the sequence of views seen during action execution.  
Furthermore, as an alternative to training the agent to make non-myopic camera motions to best reduce reconstruction error in a fixed budget of glimpses, one could instead formulate the objective in terms of a fixed threshold on reconstruction error, and allow the agent to move until that threshold is reached.   The former (our formulation) is valuable for scenarios with hard resource constraints; the latter is valuable for scenarios with hard accuracy constraints.

A third limitation of the current approach is that in practice we find that the diversity of actions selected by our learned policies is sometimes limited. The agent often tends to prefer a reduced action space of two or three actions depending on the starting point and the environment, despite using a loss term explicitly encouraging high entropy of selected actions. We believe that this could be related to optimization difficulties commonly associated with policy-gradient-based reinforcement learning, and improvements on this front would also improve the performance of our approach.

Our approach is also affected by a well-known limitation associated with rectangular representations of spherical environments~\cite{coors2018spherenet} where information at the poles are oversampled  compared to the central elevations, resulting in redundant information across different azimuths at the poles. This is further exacerbated in realistic scenes where the poles often represent the sky, floor and ceiling which tend to have limited diversity. Due to this issue, we observed that heuristic policies which sample constant actions while avoiding the poles compete strongly with learned approaches and even outperform supervised policies in some cases. We found that incorporating priors which encourage the agent to move away from the poles would result in consistent performance gains for our method as well. One future direction to avoid the issue would be to design environments that have varying azimuths across elevations.

Another drawback is that our current testbeds  handle only camera rotations, not translations. In future work, we will extend our approach to 3D environments that also permit camera translations~\cite{wu2018building,mattersim}.  In such scenarios, intelligent look-around behavior becomes even more essential, since no matter what visual sensors it has, an agent must move its camera to observe another room. We also plan to consider other tasks for transfer such as target-driven navigation~\cite{savinov2018semi} and model-based RL~\cite{ha2018world,piergiovanni2018learning}, where a preliminary exploratory stage is crucial for performing well on downstream tasks.

Finally, it will  be interesting future work to explore how multiple sensing modalities could  work together to learn look-around behavior.  For example, an agent that hears a sudden noise from one direction might learn to look there to gain new information about dynamic objects in the scene.  Or, an agent that sees an unfamiliar texture might reach out to touch the object surface to better anticipate its shape.

\FloatBarrier
\section*{Materials and Methods}

In this final section, we overview the implementation of our approach.  Complete implementation details are provided in the Supplementary Materials.

\subsection*{Recurrent observation completion network}
\label{sec:completion}

\begin{figure}[t]
  \centering
  \includegraphics[width=0.9\linewidth, trim=1cm 0 0 0, clip]{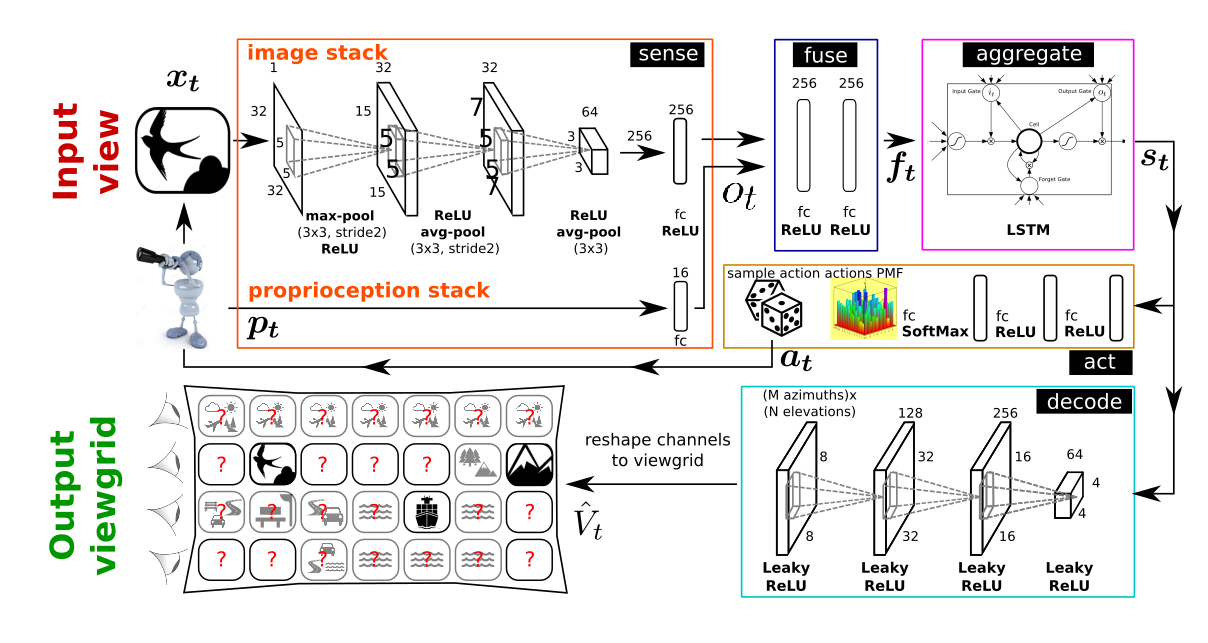} 
  \caption{Architecture of our active observation completion system. While the input-output pair shown here is for the case of $360^{\circ}$ scenes, we use the same architecture for the case of 3D objects. In the output viewgrid, solid black portions denote observed views, question marks denote unobserved views, and transparent black portions denote the system's uncertain contextual guesses.} 
  \label{fig:architecture}
\end{figure}

We now discuss the recurrent neural network used for active observation completion. The architecture naturally splits into five modules with distinct functions: \textsc{sense}, \textsc{fuse}, \textsc{aggregate}, \textsc{decode}, and \textsc{act}. 
Architecture details for all modules are given in Figure 7.

\subsubsection*{Encoding to an internal model of the target} First, we define the core modules with which the agent encodes its internal model of the current environment.  At each step $t$, the agent is presented with a 2D view $\bm{x}_t$ captured from a new viewpoint $\bm{\theta}_t$. We stress that absolute viewpoint coordinates $\bm{\theta}_t$ are not fully known, and objects/scenes are not presented in any canonical orientation. All viewgrids inferred by our approach treat the first view's azimuth as the origin.  We assume only that the absolute elevation can be sensed using gravity, and that the agent is aware of the relative motion from the previous view. Let $\bm{p}_t$ denote this proprioceptive metadata (elevation, relative motion).

The \textsc{sense} module processes these inputs in separate neural network stacks to produce two vector outputs, which we jointly denote as $\bm{o}_t=\textsc{sense}(\bm{x}_t,\bm{p}_t)$
({see Figure 7, top left}).
\textsc{fuse} combines information from both input streams and embeds it into $\bm{f}_t=\textsc{fuse}(\bm{o}_t)$ (Figure 7, top center). 
Then this combined sensory information $\bm{f}_t$ from the current observation is fed into \textsc{aggregate}, which is a long short term memory module (LSTM)~\cite{hochreiter1997long}. \textsc{aggregate} maintains an encoded internal model $\bm{s}_t$ of the object/scene under observation to ``remember'' all relevant information from past observations. At each timestep, it updates this code, combining it with the current observation to produce $\bm{s}_t=\textsc{aggregate}(\bm{f}_1,\cdots,\bm{f}_t)$ (Figure 7, top right).

\textsc{sense}, \textsc{fuse}, and \textsc{aggregate} together encode observations into an internal state $\bm{s}_t$ that is used to produce the output viewgrid and select the action, respectively, as we detail next.

\subsubsection*{Decoding to the inferred viewgrid} \textsc{decode} translates the aggregated code into the predicted viewgrid $\hat{V}_t(\bm{x}_1,\cdots,\bm{x}_t)=\textsc{decode}(\bm{s}_t)$. To do this, it first reshapes $\bm{s}_t$ into a sequence of small 2D feature maps (Figure 7, bottom right),
before upsampling to the target dimensions using a series of learned up-convolutions.  The final up-convolution produces $MN$ maps, one for each of the $MN$ views in the viewgrid. For color images, we produce $3MN$ maps, one for each color channel of each view. This is then reshaped into the target viewgrid (Figure 7, bottom center).
Seen views are pasted directly from memory.

\subsubsection*{Acting to select the next viewpoint to observe} Finally, \textsc{act} processes the aggregate code $\bm{s}_t$ to issue a motor command $\bm{a}_t=\textsc{act}(\bm{s}_t)$ (Figure 7, middle right).
For objects, the motor commands rotate the object (i.e., agent manipulates the object or peers around it); for scenes, the motor commands move the camera (i.e., agent turns in the 3D environment).  Upon execution, the {observation's pose} updates for the next timestep to $\bm{\theta}_{t+1}=\bm{\theta}_t + \bm{a}_t$. For $t=1$, $\bm{\theta}_1$ is randomly sampled, corresponding to the agent initially encountering the new environment or object from an arbitrary pose.

Internally, $\textsc{act}$ first produces a distribution over all possible actions, and then samples $\bm{a}_t$ from this distribution.
We restrict \textsc{act} to select small discrete actions at each timestep to approximately simulate continuous motion. Once the new viewpoint $\bm{\theta}_{t+1}$ is set, a new view is captured and the whole process repeats. This happens until $T$ timesteps have passed, involving $T-1$ actions. These modules are learned end-to-end in a policy learning framework as described in the section below on the policy learning formulation.

\subsection*{Sidekick policy learning}
We now describe the sidekicks used to learn faster and converge to better policies under partial observability. In order to effectively learn to perform the task, the agent has to use the limited information available from its egocentric view to (1) aggregate information, (2) select intelligent actions to improve its training, and (3) decode the entire viewgrid.  This poses significant hurdles for policy learning under partial observability, that is, making decisions while lacking full state knowledge.  For example, our agent will not know the entire 360 environment before it must decide where to look next.  In order to address these issues, we propose sidekicks that exploit full observability available exclusively during training to aid policy learning of the ultimate agent. The key idea is to solve a simpler problem with relevance to the actual look-around task using full observability, and then transfer the knowledge to the main agent.
We define two types of sidekicks, reward-based and demonstration-based.

\subsubsection*{Reward-based sidekick}
The reward-based sidekick aims to identify a set $\{x(X,\theta_{i})\}_{i=1}^{K}$ of $K$ highly informative views in the environment $X$ by exploiting full observability during training.  
It considers a simplified completion problem where the goal is evaluate the information content of individual views themselves, i.e., to identify information hotspots in the environment that strongly suggest other parts of the environment.  For example, it might learn that facing the blank ceiling of a kitchen is less informative than looking at the contents of the refrigerator or stove.

To evaluate the informativeness of a candidate view, the sidekick sees how well the
entire environment can be reconstructed given only that view.
We train a completion model 
that can reconstruct $\hat{V}(X)$ from any single view (i.e., we set $T=1$). The score assigned to a candidate view is inversely proportional to the reconstruction error of the entire environment given only that view. 
The sidekick conveys the results to the agent during policy learning in the form of an augmented reward $r_{t}^{s}$ at each time step. Please see the section on sidekick policy learning in the Supplementary for more details. 

\subsubsection*{Demonstration-based sidekick}
\label{sec:demo}
Our second sidekick generates trajectories of informative views through a sidekick policy $\pi_{s}$.  In a trajectory, the informativeness of the current view is conditioned on the past views selected, as opposed to sampling individually informative views.  To condition the informativeness on past views, we use a cumulative coverage score (see Eqn.~9, 10 in Supplementary) that measures the amount of information gathered about different parts of the environment until time $t$. 
The goodness of a view is measured by the increase in cumulative coverage obtained upon selecting that view, i.e., how well it complements the previously selected views.
Please see the section on sidekick policy learning in the supplementary material for full details.

The demonstration sidekick uses this coverage score to sample informative trajectories. 
Given a starting view in $X$, the demonstration sidekick selects a trajectory of $T$ views that are jointly maximize the coverage of $X$. 
At each time step, the demonstration sidekick evaluates the gain in cumulative coverage obtained by sampling each view in its neighborhood and then greedily samples the best view (see Eqn.~11 in Supplementary).
 
We use sidekick-generated trajectories as supervision to the agent for a short preparatory period. The goal is to initialize the agent with useful insights learned by the sidekick to accelerate training of better policies. We achieve this through a hybrid training procedure that combines imitation and reinforcement, as described in the demonstration-based sidekick section in the supplementary material.

\subsection*{Policy learning formulation}

Having defined the recurrent network model and the sidekick policy preparation, we
now describe the policy learning framework used to train our agent as well as the mechanisms used to incorporate sidekick rewards ($r_{t}^{s}$) and demonstrations (obtained from $\pi_{s}$).  All modules are jointly optimized end-to-end to improve the final reconstructed viewgrid $\hat{V}_T$, which contains predicted views $\bm{\hat{x}_T}(X,\bm{\theta}_j)$ for all viewpoints $\bm{\theta}_j, 1\leq j \leq MN$. 
The agent learns a policy $\pi(a | s_t)$ which returns a distribution over actions for the aggregated internal representation 
$s_t$ at time $t$. Let $\mathcal{A} = \{a_i\}$ denote the set of camera motions available to the agent. Our agent seeks the policy that minimizes reconstruction error for the environment given a budget of $T$ camera motions (views). Let $W_{s}, W_{f}, W_{r}, W_{d}, W_{a}$ represent the weights of the \textsc{Sense}, \textsc{Fuse}, \textsc{Aggregate}, \textsc{Decode}, and \textsc{Act} modules. If we denote the set of weights of the network $[W_{s}, W_{f}, W_{r}, W_{d}, W_{a}]$ by $W$, and $W$ excluding $W_{a}$ by $W_{/a}$, and $W$ excluding $W_{d}$ by $W_{/d}$, then  the overall weight update is: 
\begin{equation}
\label{eqn:basic_weight_update}
\Delta W = \frac{1}{n} \sum_{j=1}^{n} \lambda_{r} \Delta W^{rec}_{/a} + \lambda_{a} \Delta W^{act}_{/d},
\end{equation}
where $n$ is the number of training samples, $j$ indexes over the training samples, $\lambda_r$ and $\lambda_a$ are constants and $\Delta W^{rec}_{/a}$ and $\Delta W^{act}_{/d}$ update all parameters except $W_{a}$ and $W_{d}$, respectively.

The pixel-wise MSE reconstruction loss ($\mathcal{L}^{rec}_{t}$) and corresponding weight update at time $t$ are as follows:
\begin{equation}
\label{rec_loss}
\begin{split}
  \mathcal{L}_{rec}^t(X) =  \sum_{i=1}^{MN} d\left(\hat{x}_{t}(X, \theta^{(i)}+\Delta_{0}), x(X, \theta^{(i)})\right), \\
  \Delta W^{rec}_{/a} = -\sum_{t=1}^{T} \nabla_{W_{/a}} \mathcal{L}_{rec}^{t}(X),
\end{split}
\end{equation}
where $\hat{x}_{t}(X, \theta^{(i)})$ denotes the reconstructed view at viewpoint $\theta^{(i)}$ and time $t$, $d$ denotes the pixelwise reconstruction MSE, and $\Delta_{0}$ denotes the offset to account for the unknown starting azimuth~\cite{dinesh-cvpr2018}.

The agent's reward at time $t$ consists of the intrinsic reward from the sidekick $r^{s}_t = \text{Info}(x(X,\theta_t),X)$ and the negated final reconstruction loss, $-\mathcal{L}_{rec}^T(X)$:
\begin{equation}
\label{rec_reward}
r_{t} = \begin{cases}
          r^{s}_{t} &\quad 1 \le t \le T-2\\
         
          -\mathcal{L}_{rec}^T(X) + r^{s}_{t} &\quad t = T-1.\\
        \end{cases}
\end{equation}
The sidekick reward $r_{t}^s$ serves to densify the rewards by exploiting full observability, thereby reducing uncertainty during policy learning. Please see Supplementary for the exact form of $r_{t}^s$.

The update from the policy consists of an actor-critic update, with a baseline $b$ to reduce variance, and supervision from the demonstration sidekick:
\begin{equation}
\label{eqn:reinforce}
\Delta W_{/d}^{act} = \sum_{t=1}^{T-1} \nabla_{W_{/d}} \text{log}\,\pi(a_{t}|s_{t})\bigg(\sum_{t^{'}=t}^{T-1}r_{t^{'}} - b(s_{t})\bigg) + \Delta W_{/d}^{demo}.\vspace{-0.10cm}
\end{equation}
We adopt the baseline $b$ as the value function from an actor-critic~\cite{sutton1998reinforcement} method to update the \textsc{Act} module.  The demonstration sidekick's supervision is defined below in Eqn.~\ref{supervised_loss}.  The \textsc{act} term additionally includes a loss to update the learned value network and entropy regularization to promote diversity in action selection (please see additional loss functions in Supplementary).

\noindent Whereas the reward sidekick augments rewards, the demonstration sidekick instead influences  policy learning by directly supervising the early rounds of action selection. This is achieved through a cross entropy loss between the sidekick's policy $\pi_s$ and the agent's policy $\pi$:\vspace{-0.10cm}
\begin{equation}
\label{supervised_loss}
\Delta W_{/d}^{demo} = \sum_{t=1}^{T-1} \sum_{a \in \mathcal{A}} \nabla_{/d}\big(\pi_s(a | s_{t})~ \text{log}\,\pi(a | s_{t})\big).\vspace{-0.10cm}
\end{equation}
Please see the sidekick policy learning section in Supplementary for the exact form of $\pi_s$.

\noindent We pretrain the \textsc{Sense}, \textsc{Fuse}, and \textsc{Decode} modules
with $T=1$. The full network is then trained end-to-end (with \textsc{Sense} and \textsc{Fuse} frozen). For training with sidekicks, the agent is augmented either with additional rewards from the reward sidekick (Eqn.~\ref{rec_reward}) or an additional supervised loss from the demonstration sidekick (Eqn.~\ref{supervised_loss}). 

\subsection*{Unsupervised policy transfer to unseen tasks}
We now describe the mechanism used to transfer policies learned in an unsupervised fashion via active observation completion to new perception tasks requiring sequential observations.  This section details the process overviewed above in the look-around policy transfer section. 
The main idea is to inject our generic look-around policy into new unseen tasks in unseen environments.  In particular, we consider transferring our policy---trained with neither manual supervision nor task-specific reward---into various semantic and geometric recognition tasks for which the agent was not specifically trained.  Recall, we consider four different tasks:
recognition, surface area estimation, light source localization, and camera pose estimation.

At training time, we train an end-to-end task-specific model (model A) with a random policy (\texttt{\small rnd-actions}), and an active observation completion model (model B). Note that  our  completion model is trained without supervision to look around environments that have zero overlap with model A's test set.  Furthermore, even the categories seen during training may differ from those during testing.  For example, the agent might see various furniture categories during training (bookcase, bed, etc.), but never a chair, yet it must generalize well to look around a chair.

At test time, both the task-specific model A and the active observation model B receive and process the same inputs at each timestep. The task-specific model does not have a learned policy of its own, since it was trained with a policy that samples random actions. At each timestep, model B selects actions to complete its internal model of the new environment based on its look-around policy. This action is then communicated to model A in place of the random-actions with which it was trained.
Therefore, model A gathers its information based on the actions provided by model B. Model A now makes a prediction for the target task. If the policy learned in model B is truly generic, it will intelligently explore to solve the new (unseen) tasks despite never receiving task-specific reward for any one of them during training. 

\pagebreak
\section*{\Huge{List of Supplementary Materials}}

\subsection*{The supplementary PDF file includes:\vspace{-0.5em}}
Text to augment the implementation details in Materials and Methods\\
\noindent Figure S1. Sidekick Framework\\
\noindent Figure S2. Light source localization example \\
\noindent Figure S3. Convergence of sidekick policy learning\\
\noindent Figure S4. Training on different target budgets $T$\\
\noindent Figure S5. Episodes of active observation completion\\
\noindent Figure S6. GAN refinement\\

\subsection*{Other Supplementary Material for this manuscript include:\vspace{-0.5em}}
\noindent Movie S1 (.mp4 format). Sample walkthroughs in reconstructed environments\\
\noindent Movie S2 (.mp4 format). Active observation completion on SUN360\\
\noindent Movie S3 (.mp4 format). Active observation completion on ModelNet\\

\noindent The materials can be found here: {\footnotesize \url{http://vision.cs.utexas.edu/projects/visual-exploration/}}
\pagebreak

\FloatBarrier
\bibliographystyle{unsrt}

\begin{thebibliography}{10}

\bibitem{ILSVRC15}
Olga Russakovsky, Jia Deng, Hao Su, Jonathan Krause, Sanjeev Satheesh, Sean Ma,
  Zhiheng Huang, Andrej Karpathy, Aditya Khosla, Michael Bernstein,
  Alexander~C. Berg, and Li~Fei-Fei.
\newblock {ImageNet Large Scale Visual Recognition Challenge}.
\newblock {\em International Journal of Computer Vision}, 2015.

\bibitem{lin2014microsoft}
Tsung-Yi Lin, Michael Maire, Serge Belongie, James Hays, Pietro Perona, Deva
  Ramanan, Piotr Doll{\'a}r, and C~Lawrence Zitnick.
\newblock Microsoft coco: Common objects in context.
\newblock In {\em European Conference on Computer Vision}, 2014.

\bibitem{soomro2012ucf101}
Khurram Soomro, Amir~Roshan Zamir, and Mubarak Shah.
\newblock Ucf101: A dataset of 101 human actions classes from videos in the
  wild.
\newblock {\em arXiv preprint arXiv:1212.0402}, 2012.

\bibitem{soska2008development}
Kasey~C Soska and Scott~P Johnson.
\newblock Development of three-dimensional object completion in infancy.
\newblock In {\em Child development}, 2008.

\bibitem{soska2010systems}
Kasey~C Soska, Karen~E Adolph, and Scott~P Johnson.
\newblock Systems in development: motor skill acquisition facilitates
  three-dimensional object completion.
\newblock In {\em Developmental psychology}, 2010.

\bibitem{kellman1983perception}
Philip~J Kellman and Elizabeth~S Spelke.
\newblock Perception of partly occluded objects in infancy.
\newblock In {\em Cognitive psychology}, 1983.

\bibitem{torralba2006contextual}
Antonio Torralba, Aude Oliva, Monica~S Castelhano, and John~M Henderson.
\newblock Contextual guidance of eye movements and attention in real-world
  scenes: the role of global features in object search.
\newblock In {\em Psychological review}, 2006.

\bibitem{dinesh-eccv2016}
Dinesh Jayaraman and Kristen Grauman.
\newblock Look-ahead before you leap: end-to-end active recognition by
  forecasting the effect of motion.
\newblock In {\em European Conference on Computer Vision}, 2016.

\bibitem{germs-bmvc2015}
Mohsen Malmir, Karan Sikka, Deborah Forster, Javier~R Movellan, and Garison
  Cottrell.
\newblock Deep q-learning for active recognition of germs: Baseline performance
  on a standardized dataset for active learning.
\newblock In {\em British Machine Vision Conference}, 2015.

\bibitem{shapenet}
Zhirong Wu, Shuran Song, Aditya Khosla, Fisher Yu, Linguang Zhang, Xiaoou Tang,
  and Jianxiong Xiao.
\newblock 3d shapenets: A deep representation for volumetric shapes.
\newblock In {\em Computer Vision and Pattern Recognition, IEEE Conference on},
  2015.

\bibitem{ammirato-icra2017}
Phil Ammirato, Patrick Poirson, Eunbyung Park, Jana Ko{\v{s}}eck{\'a}, and
  Alexander~C Berg.
\newblock A dataset for developing and benchmarking active vision.
\newblock In {\em Robotics and Automation, IEEE International Conference on},
  2017.

\bibitem{yeung2016end}
Serena Yeung, Olga Russakovsky, Greg Mori, and Li~Fei-Fei.
\newblock End-to-end learning of action detection from frame glimpses in
  videos.
\newblock In {\em Computer Vision and Pattern Recognition, IEEE Conference on},
  2016.

\bibitem{mathe-cvpr2016}
S.~Mathe, A.~Pirinen, and C.~Sminchisescu.
\newblock Reinforcement learning for visual object detection.
\newblock In {\em Computer Vision and Pattern Recognition, IEEE Conference on},
  2016.

\bibitem{timely-nips2012}
S.~Karayev, T.~Baumgartner, M.~Fritz, and T.~Darrell.
\newblock Timely object recognition.
\newblock In {\em Advances in Neural Information Processing Systems}, 2012.

\bibitem{pathakICMl17curiosity}
Deepak Pathak, Pulkit Agrawal, Alexei~A. Efros, and Trevor Darrell.
\newblock Curiosity-driven exploration by self-supervised prediction.
\newblock In {\em International Conference on Machine Learning}, 2017.

\bibitem{chen2018learning}
Tao Chen, Saurabh Gupta, and Abhinav Gupta.
\newblock Learning exploration policies for navigation.
\newblock In {\em International Conference on Learning Representations}, 2019.

\bibitem{Hepp_2018_ECCV}
Benjamin Hepp, Debadeepta Dey, Sudipta~N. Sinha, Ashish Kapoor, Neel Joshi, and
  Otmar Hilliges.
\newblock Learn-to-score: Efficient 3d scene exploration by predicting view
  utility.
\newblock In {\em The European Conference on Computer Vision}, September 2018.

\bibitem{song2018im2pano3d}
Shuran Song, Andy Zeng, Angel~X Chang, Manolis Savva, Silvio Savarese, and
  Thomas Funkhouser.
\newblock Im2pano3d: Extrapolating 360° structure and semantics beyond the
  field of view.
\newblock In {\em Computer Vision and Pattern Recognition, IEEE Conference on},
  pages 3847--3856, 2018.

\bibitem{ji-cvpr2017}
Dinghuang Ji, Junghyun Kwon, Max McFarland, and Silvio Savarese.
\newblock Deep view morphing.
\newblock In {\em Computer Vision and Pattern Recognition, IEEE Conference on},
  volume~2, 2017.

\bibitem{kulkarni-nips2015}
Tejas~D Kulkarni, William~F Whitney, Pushmeet Kohli, and Josh Tenenbaum.
\newblock Deep convolutional inverse graphics network.
\newblock In {\em Advances in neural information processing systems}, pages
  2539--2547, 2015.

\bibitem{jayaraman-eccv2018}
Dinesh Jayaraman, Ruohan Gao, and Kristen Grauman.
\newblock Shapecodes: Self-supervised feature learning by lifting views to
  viewgrids.
\newblock {\em European Conference on Computer Vision}, 2018.

\bibitem{gqn-science2018}
SM~Ali Eslami, Danilo~Jimenez Rezende, Frederic Besse, Fabio Viola, Ari~S
  Morcos, Marta Garnelo, Avraham Ruderman, Andrei~A Rusu, Ivo Danihelka, Karol
  Gregor, et~al.
\newblock Neural scene representation and rendering.
\newblock {\em Science}, 360(6394):1204--1210, 2018.

\bibitem{dinesh-cvpr2018}
Dinesh Jayaraman and Kristen Grauman.
\newblock Learning to look around: Intelligently exploring unseen environments
  for unknown tasks.
\newblock In {\em Computer Vision and Pattern Recognition, IEEE Conference on},
  2018.

\bibitem{santhosh-eccv2018}
Santhosh~K. Ramakrishnan and Kristen Grauman.
\newblock {Sidekick Policy Learning for Active Visual Exploration}.
\newblock In {\em European Conference on Computer Vision}, 2018.

\bibitem{johns-cvpr2016}
Edward Johns, Stefan Leutenegger, and Andrew~J Davison.
\newblock Pairwise decomposition of image sequences for active multi-view
  recognition.
\newblock In {\em Computer Vision and Pattern Recognition, IEEE Conference on},
  2016.

\bibitem{zhu-iccv2017}
Yuke Zhu, Daniel Gordon, Eric Kolve, Dieter Fox, Li~Fei-Fei, Abhinav Gupta,
  Roozbeh Mottaghi, and Ali Farhadi.
\newblock {Visual Semantic Planning using Deep Successor Representations}.
\newblock In {\em Computer Vision, IEEE International Conference on}, 2017.

\bibitem{gupta2017unifying}
Saurabh Gupta, David Fouhey, Sergey Levine, and Jitendra Malik.
\newblock Unifying map and landmark based representations for visual
  navigation.
\newblock {\em arXiv preprint arXiv:1712.08125}, 2017.

\bibitem{zhu2017target}
Yuke Zhu, Roozbeh Mottaghi, Eric Kolve, Joseph~J Lim, Abhinav Gupta,
  Li~Fei-Fei, and Ali Farhadi.
\newblock Target-driven visual navigation in indoor scenes using deep
  reinforcement learning.
\newblock In {\em Robotics and Automation, IEEE International Conference on},
  2017.

\bibitem{dinesh-pami2018}
D.~Jayaraman and K.~Grauman.
\newblock End-to-end policy learning for active visual categorization.
\newblock {\em IEEE Transactions on Pattern Analysis and Machine Intelligence},
  2018.

\bibitem{guo2014deep}
Xiaoxiao Guo, Satinder Singh, Honglak Lee, Richard~L Lewis, and Xiaoshi Wang.
\newblock Deep learning for real-time atari game play using offline monte-carlo
  tree search planning.
\newblock In {\em Advances in Neural Information Processing Systems}, 2014.

\bibitem{vapnik2016learning}
Vladimir Vapnik and Rauf Izmailov.
\newblock Learning with intelligent teacher.
\newblock In {\em Symposium on Conformal and Probabilistic Prediction with
  Applications}, 2016.

\bibitem{xiao2012recognizing}
Jianxiong Xiao, Krista~A Ehinger, Aude Oliva, and Antonio Torralba.
\newblock Recognizing scene viewpoint using panoramic place representation.
\newblock In {\em Computer Vision and Pattern Recognition, IEEE Conference on},
  2012.

\bibitem{harel2006graph}
Jonathan Harel, Christof Koch, and Pietro Perona.
\newblock Graph-based visual saliency.
\newblock In {\em Advances in Neural Information Processing Systems}, 2006.

\bibitem{isola2017image}
Phillip Isola, Jun-Yan Zhu, Tinghui Zhou, and Alexei~A Efros.
\newblock Image-to-image translation with conditional adversarial networks.
\newblock In {\em Computer Vision and Pattern Recognition, IEEE Conference on},
  pages 5967--5976. IEEE, 2017.

\bibitem{choy20163d}
Christopher~B Choy, Danfei Xu, JunYoung Gwak, Kevin Chen, and Silvio Savarese.
\newblock 3d-r2n2: A unified approach for single and multi-view 3d object
  reconstruction.
\newblock In {\em Proceedings of the European Conference on Computer Vision
  ({ECCV})}, 2016.

\bibitem{Fan_2017_CVPR}
Haoqiang Fan, Hao Su, and Leonidas~J. Guibas.
\newblock A point set generation network for 3d object reconstruction from a
  single image.
\newblock In {\em Computer Vision and Pattern Recognition, IEEE Conference on},
  July 2017.

\bibitem{wang2018pixel2mesh}
Nanyang Wang, Yinda Zhang, Zhuwen Li, Yanwei Fu, Wei Liu, and Yu-Gang Jiang.
\newblock Pixel2mesh: Generating 3d mesh models from single rgb images.
\newblock {\em arXiv preprint arXiv:1804.01654}, 2018.

\bibitem{dosovitskiy2017carla}
Alexey Dosovitskiy, German Ros, Felipe Codevilla, Antonio Lopez, and Vladlen
  Koltun.
\newblock Carla: An open urban driving simulator.
\newblock In {\em Conference on Robot Learning}, 2017.

\bibitem{pinto2017asymmetric}
Lerrel Pinto, Marcin Andrychowicz, Peter Welinder, Wojciech Zaremba, and Pieter
  Abbeel.
\newblock Asymmetric actor critic for image-based robot learning.
\newblock {\em Robotics: Science and Systems}, 2018.

\bibitem{embodiedqa}
Abhishek Das, Samyak Datta, Georgia Gkioxari, Stefan Lee, Devi Parikh, and
  Dhruv Batra.
\newblock {E}mbodied {Q}uestion {A}nswering.
\newblock In {\em Computer Vision and Pattern Recognition, IEEE Conference on},
  2018.

\bibitem{wu2018building}
Yi~Wu, Yuxin Wu, Georgia Gkioxari, and Yuandong Tian.
\newblock Building generalizable agents with a realistic and rich 3d
  environment.
\newblock {\em arXiv preprint arXiv:1801.02209}, 2018.

\bibitem{mattersim}
Peter Anderson, Qi~Wu, Damien Teney, Jake Bruce, Mark Johnson, Niko
  S{\"u}nderhauf, Ian Reid, Stephen Gould, and Anton van~den Hengel.
\newblock Vision-and-language navigation: Interpreting visually-grounded
  navigation instructions in real environments.
\newblock In {\em Computer Vision and Pattern Recognition, IEEE Conference on},
  2018.

\bibitem{savinov2018semi}
Nikolay Savinov, Alexey Dosovitskiy, and Vladlen Koltun.
\newblock Semi-parametric topological memory for navigation.
\newblock {\em International Conference on Learning Representations}, 2018.

\bibitem{ha2018world}
David Ha and J{\"u}rgen Schmidhuber.
\newblock World models.
\newblock {\em arXiv preprint arXiv:1803.10122}, 2018.

\bibitem{piergiovanni2018learning}
AJ~Piergiovanni, Alan Wu, and Michael~S Ryoo.
\newblock Learning real-world robot policies by dreaming.
\newblock {\em arXiv preprint arXiv:1805.07813}, 2018.

\bibitem{hochreiter1997long}
Sepp Hochreiter and J{\"u}rgen Schmidhuber.
\newblock Long short-term memory.
\newblock {\em Neural computation}, 9(8):1735--1780, 1997.

\bibitem{neal1990learning}
Radford~M Neal.
\newblock Learning stochastic feedforward networks.
\newblock {\em Department of Computer Science, University of Toronto}, 64(9),
  1990.

\bibitem{sutton1998reinforcement}
Richard~S Sutton and Andrew~G Barto.
\newblock {\em Reinforcement learning: An introduction}.

\bibitem{bojarski2016end}
Mariusz Bojarski, Davide Del~Testa, Daniel Dworakowski, Bernhard Firner, Beat
  Flepp, Prasoon Goyal, Lawrence~D Jackel, Mathew Monfort, Urs Muller, Jiakai
  Zhang, et~al.
\newblock End to end learning for self-driving cars.
\newblock {\em arXiv preprint arXiv:1604.07316}, 2016.

\bibitem{giusti2016machine}
Alessandro Giusti, J{\'e}r{\^o}me Guzzi, Dan~C Cire{\c{s}}an, Fang-Lin He,
  Juan~P Rodr{\'\i}guez, Flavio Fontana, Matthias Faessler, Christian Forster,
  J{\"u}rgen Schmidhuber, Gianni Di~Caro, et~al.
\newblock A machine learning approach to visual perception of forest trails for
  mobile robots.
\newblock {\em IEEE Robotics and Automation Letters}, 2016.

\bibitem{duan2017one}
Yan Duan, Marcin Andrychowicz, Bradly Stadie, OpenAI~Jonathan Ho, Jonas
  Schneider, Ilya Sutskever, Pieter Abbeel, and Wojciech Zaremba.
\newblock One-shot imitation learning.
\newblock In {\em Advances in Neural Information Processing Systems}, 2017.

\bibitem{goodfellow2014generative}
Ian Goodfellow, Jean Pouget-Abadie, Mehdi Mirza, Bing Xu, David Warde-Farley,
  Sherjil Ozair, Aaron Courville, and Yoshua Bengio.
\newblock Generative adversarial nets.
\newblock In {\em Advances in neural information processing systems}, pages
  2672--2680, 2014.

\bibitem{coors2018spherenet}
Coors, Benjamin and Paul Condurache, Alexandru and Geiger, Andreas.
\newblock Spherenet: Learning spherical representations for detection and classification in omnidirectional images.
\newblock In {\em Proceedings of the European Conference on Computer Vision (ECCV)}, 2018.

\bibitem{footnote1}
For simplicity of presentation, we represent an ``environment'' as $X$ where the agent explores a novel 
 scene, looking outward in new viewing directions.  However, experiments will  also use $X$ as an object where the agent moves around an object, 
 looking inward at it from new viewing angles. 
 Figure 1 illustrates the two scenarios.

\bibitem{footnote2}
The angles were selected to break symmetry and reduce redundancy of views.

\bibitem{footnote3}
For the sake of brevity, we report the best performances among the two sidekick variants we proposed in~\cite{santhosh-eccv2018}.

\bibitem{footnote4}
We refine the decoded viewgrids (for both our method and the baseline) with a pix2pix~\cite{isola2017image}-style conditional Generative Adversarial Network (GAN), detailed in the Supplementary Materials.

\end{thebibliography}

\noindent\textbf{Acknowledgements:} We thank Yu-Chuan Su, Kimberly Hsiao, Bo Xiong and Philipp Kr\"ahenb\"uhl for helpful discussions. 

\noindent\textbf{Funding:} The University of Texas at AUstin is supported in part by DARPA Lifelong Learning Machines (L2M), an AFOSR DURIP equipment grant, an IBM OCR Award, and a Sony Research Award. This material is based on research sponsored by the Air Force Research Laboratory and DARPA under agreement number FA8750-18-2-0126. The U.S. Government is authorized to reproduce and distribute reprints for Governmental purposes notwithstanding any copyright notation thereon. 

\noindent\textbf{Author contributions:} The project ideas were conceived by D.J, S.R and K.G. The experiments were designed by D.J, S.R and K.G. The experiments were performed by D.J and S.R. The paper was written by D.J, S.R and K.G.

\noindent\textbf{Competing interests:} The authors declare that they have no competing interests. 

\noindent\textbf{Data and materials availability:} The datasets for this study are available at {\small\url{http://vision.cs.utexas.edu/projects/sidekicks/scirobo-2019-data.zip}} and the source code is available at {\small\url{ https://github.com/srama2512/sidekicks}}.

\FloatBarrier
\pagebreak

\section*{\Huge{Supplementary Materials}}

\setcounter{figure}{0}
\renewcommand{\thefigure}{S\arabic{figure}}

\section*{Sidekick Policy Learning}
We now describe the exact forms of the sidekick reward $r_{t}^{s}$ and the sidekick policy $\pi_{s}$ learned by the reward-based and demonstration-based sidekicks, respectively.
\begin{figure}[t]
    \centering
      \includegraphics[width=\textwidth, clip, trim={0cm 0cm 0cm 0cm}]{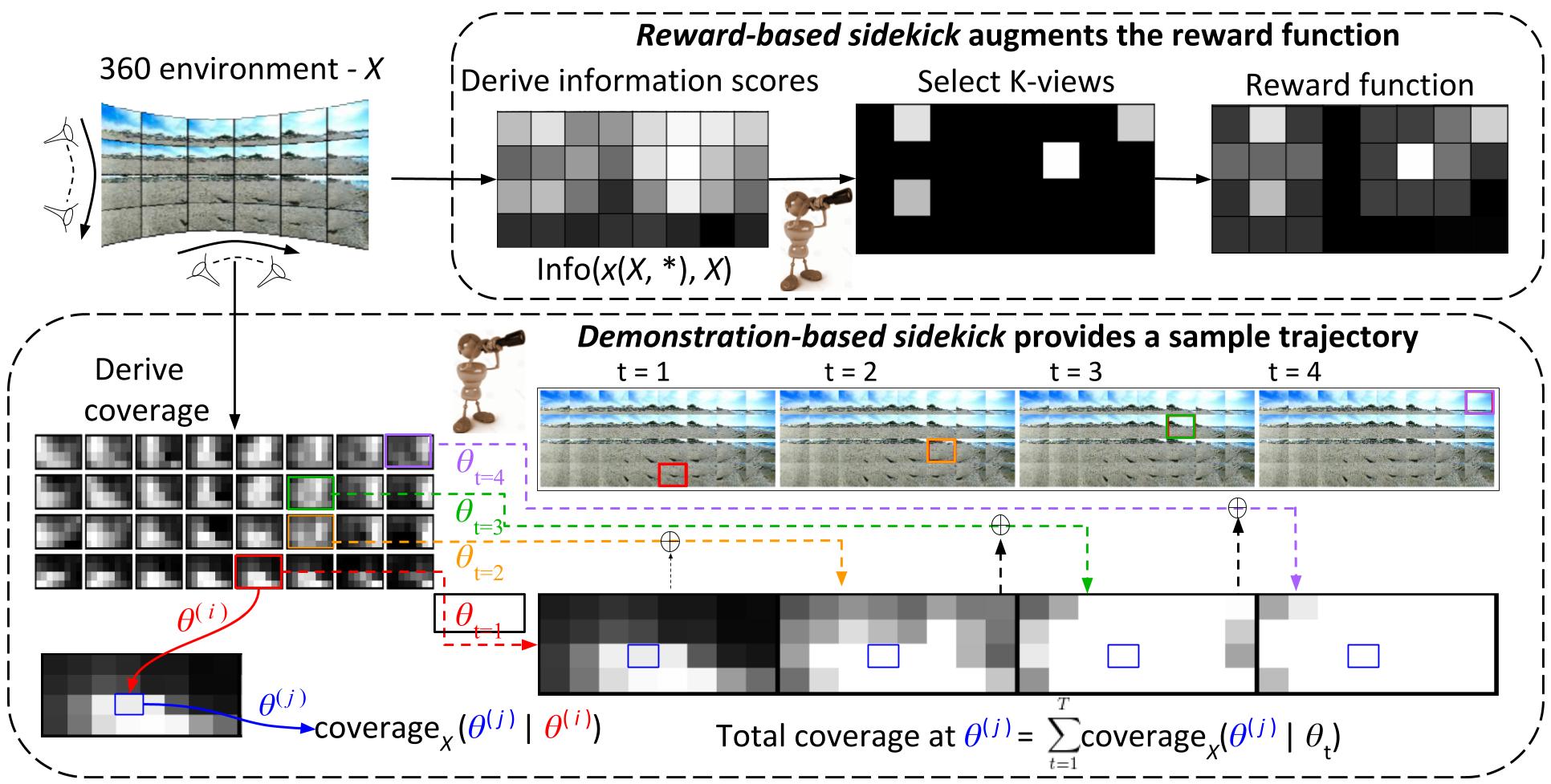}

    \caption{Sidekick Framework: Top left shows the $360^{\circ}$ environment's viewgrid, indexed by viewing elevation and azimuth.
    Top: Reward sidekick --- 
 scores individual views based on how well they alone permit inference of the viewgrid $X$ (Eq~\ref{one_view_score}).  The grid of scores (center) is post-processed with non-max suppression to prioritize $K$ non-redundant views (right), then is used to shape the agent's rewards.
    Bottom: Demonstration sidekick --- Left ``grid-of-grids" displays example coverage score maps (Eq~\ref{coverage}) for all $\theta^{(i)},\theta^{(j)}$ view pairs.  The outer $N \times M$ grid considers each  $\theta^{(i)}$, and each inner $N \times M$ grid considers each $\theta^{(j)}$ for the given $\theta^{(i)}$ (bottom left).  A pixel in that grid is bright if coverage is high for $\theta^{(j)}$ given $\theta^{(i)}$, and dark otherwise.  Each $\theta^{(i)}$ denotes an (elevation, azimuth) pair.  
While observed views and their neighbors are naturally recoverable (brighter), the sidekick uses broader environment context to also anticipate distant and/or different-looking parts of the environment, as seen by the non-uniform spread of scores in the left grid.
    Given the coverage function and a starting position, this sidekick selects actions to greedily optimize the coverage objective (Eq~\ref{coverage_objective}).
    The bottom right strip shows the cumulative coverage maps as each of the $T$=4 glimpses is selected.}
    \label{sidekick-framework}
\end{figure}

\subsection*{Reward-based sidekick}
As mentioned in Materials and Methods, the score assigned to a candidate view is inversely proportional to the reconstruction error of the entire environment given only that view. Here, we elaborate on how this computation is performed. Let $\hat{V}(X | y)$ denote the decoded reconstruction for $X$ given only view $y$ as input. This is obtained from the \texttt{\small one-view} model which was originally presented as a baseline in Results.
The sidekick scores the information in observation $x(X, \theta)$ as:
\begin{equation}
  \label{one_view_score}
  \text{Info}\left(x(X, \theta), X\right) ~~\propto^{-1}~~d\left(\hat{V}(X|x(X, \theta)), V(X)\right),
\end{equation}
where $d$ denotes the reconstruction error and $V(X)$ is the fully observed environment.  We use a simple pixelwise reconstruction MSE loss for $d$ to quantify information. 
Higher-level losses, e.g., for detected objects, could be employed when available; pixel loss is most general in that it avoids committing to any particular label space or task.
The scores are normalized to lie in $[0, 1]$ across the different views of $X$. 
The sidekick scores each candidate view.  
Then, in order to sharpen the effects of the scoring function and avoid favoring redundant observations, the sidekick selects the top $K$ most informative views with greedy non-maximal suppression.  It iteratively selects
the view with the highest score and suppresses all views in the neighborhood of that view until $K$ views are selected.  

This computation yields a map of favored views for each training environment.  See Figure S1, top row. The map shows the sidekick reward $r_{t}^{s}$ that is provided to the agent for visiting each view in the environment, i.e., $r_{t}^{s} = \text{Info}(x(X, \theta_{t}), X)$ where $\theta_{t}$ is the view visited by the agent at time $t$. Note that while the sidekick indexes views in absolute angles, the agent will not; all its observations are relative to its initial (random) glimpse direction.  This works because the sidekick becomes a part of the environment, meaning it attaches rewards to the true views of the environment.  
In short, the reward-based sidekick shapes rewards based on its exploration with full observability.

\subsection*{Demonstration-based sidekick}
As mentioned in Materials and Methods, the demonstration sidekick selects a trajectory of $T$ views that are deemed to be most informative about the environment $X$.  In contrast to the reward-based sidekick, the informativeness of a view is conditioned on the previously selected views and is quantified using coverage. Here, we describe the formulation of coverage used, greedy view sampling, and the subsequent supervision provided to the main agent.

Coverage reflects how much information $x(X, \theta)$ contains about all other views in $X$. The coverage score for view $\theta^{(j)}$ upon selecting view $\theta^{(i)}$ is:
\begin{equation}
  \label{coverage}
  \text{Coverage}_{X}\left(\theta^{(j)} | \theta^{(i)}\right) \propto^{-1} d\left(\hat{x}(X, \theta^{(j)}), x(X, \theta^{(j)}) \right),
\end{equation}
where $\hat{x}$ denotes an inferred view within $\hat{V}(X | x(X, \theta^{(i)}))$, as estimated using the same $T=1$ completion network used by the reward-based sidekick, and $d$ is again the MSE loss function computed using $\ell_{2}$ distance.
Coverage scores are normalized to lie in $[0,1]$ 
for  $ 1 \le i, j \le MN$:\vspace*{-0.10in}  
\begin{equation}
  \label{coverage_objective}
  \mathcal{C}(\Theta, X) = \sum_{j=1}^{MN} \sum_{\theta\in\Theta} \text{Coverage}_{X}(\theta^{(j)} | \theta).
\end{equation}
The goal of the demonstration sidekick is to maximize the coverage objective (Eqn.~\ref{coverage_objective}), where $\Theta  = \{\theta_{1}, \ldots, \theta_{t}\}$ denotes the sequence of selected views, and $\mathcal{C}(\Theta, X)$ saturates at 1.
In other words, it seeks a sequence of reachable views such that all environment views are explained as well as possible.  See Figure S1, bottom panel. 

The policy of the sidekick ($\pi_{s}$) is to greedily select actions based on the coverage objective.  The objective encourages the sidekick to select views such that the overall
information obtained about each view in $X$ is maximized. At each time step, the sidekick looks at all available actions $a \in \mathcal{A}$ and samples the action that leads to maximum increase in coverage $C(., X)$:  
\begin{equation}
  \label{expert_policy}
  \pi_{s}(\Theta) = \argmax_{a}~\mathcal{C}\left(\Theta \cup \{\theta_{t} + a\}, X\right).
\end{equation}

In order to aid the main agent's training, we provide the generated trajectories as supervision. We achieve this through a hybrid training procedure that combines imitation and reinforcement. In particular, for the first $t_{sup}$ time steps, we let the sidekick drive the action selection and train the policy based on that supervised objective. For steps $t_{sup}$ to $T$, we let the agent's policy drive the action selection and use actor-critic~\cite{sutton1998reinforcement} to update the agent's policy (more on this in the next section).  We start with $t_{sup} = T$ and gradually reduce it to $0$ in the preparatory sidekick phase (reduction of 1 after every 50 epochs of training).
This step relates to behavior cloning~\cite{bojarski2016end,giusti2016machine,duan2017one}, which formulates policy learning as supervised action classification given states.  However, unlike typical behavior cloning, the sidekick is not an expert.  It solves a simpler version of the task, then backs away as the agent takes over to train with partial observability.

\section*{Implementation details}

Key notations:
\begin{itemize}
  \item{$p_{t}$ - proprioception input. It consists of the relative change in elevation and azimuth from $t-1$ to $t$ together with the absolute elevation at $t$. }
  \item{$x_{t}$ - input view.  Its dimensionality is $C\times H \times W$, where $C$ is the number of channels, $H$ is the image height, and $W$ is the image width. For SUN360, $C=3, H=32, W=32$ (color images, three channels), and for ModelNet, $C=1, H=32, W=32$ (grayscale images, one channel).} 
  \item{$M$ - number of azimuths in $X$ ($8$ for SUN360 and $10$ for ModelNet).}
  \item{$N$ - number of elevations in $X$ ($4$ for SUN360 and $6$ for ModelNet).}
  \item{$a_{t}$ - action taken at time $t$.}
\end{itemize}

We build on the publicly available implementation from~\cite{santhosh-eccv2018} in PyTorch. We use a one-layer recurrent neural network (RNN) with the hidden state size fixed to 256. We use the Adam optimizer with a learning rate of $0.0001 - 0.003$, weight decay of 1e-6, and other default settings from PyTorch~\footnote{Please refer to \texttt{http://pytorch.org/docs/master/optim.html}}. We set $\lambda_{r}=1$ and $\lambda_{a}=1$ based on grid search (see Eqn.~\ref{eqn:basic_weight_update}). All models are trained with three different random seeds and the results are averaged over them. In the case of the demonstration-based sidekick, we decay $T_{sup}$ from $T-1$ to $0$ after every 50 epochs. For the reward-based sidekick, we decay the rewards by a factor of $1-2$ after every $100-500$ epochs (selected based on grid search). All the models are trained for $1000$ epochs. For the reward-based sidekick, we use a non-maximal suppression neighborhood of $1$ and $K = 4$ views for SUN360, and a neighborhood of $2$ and $K = 6$ views for ModelNet.   The neighborhood and number of views were selected manually upon brief visual inspection of a few reconstructed viewgrids for each dataset to ensure sufficient spread of rewards on the viewgrid. All code and datasets will be made available.

\section*{Light source localization example}
As we explain in the main paper, light source localization is posed as a four-way classification problem where each class is identified by the elevation of the light source. The azimuth of the light source is varied uniformly randomly for each class. Figure S2 illustrates this process with an example. 
\begin{figure}
    \centering
    \includegraphics[width=\textwidth]{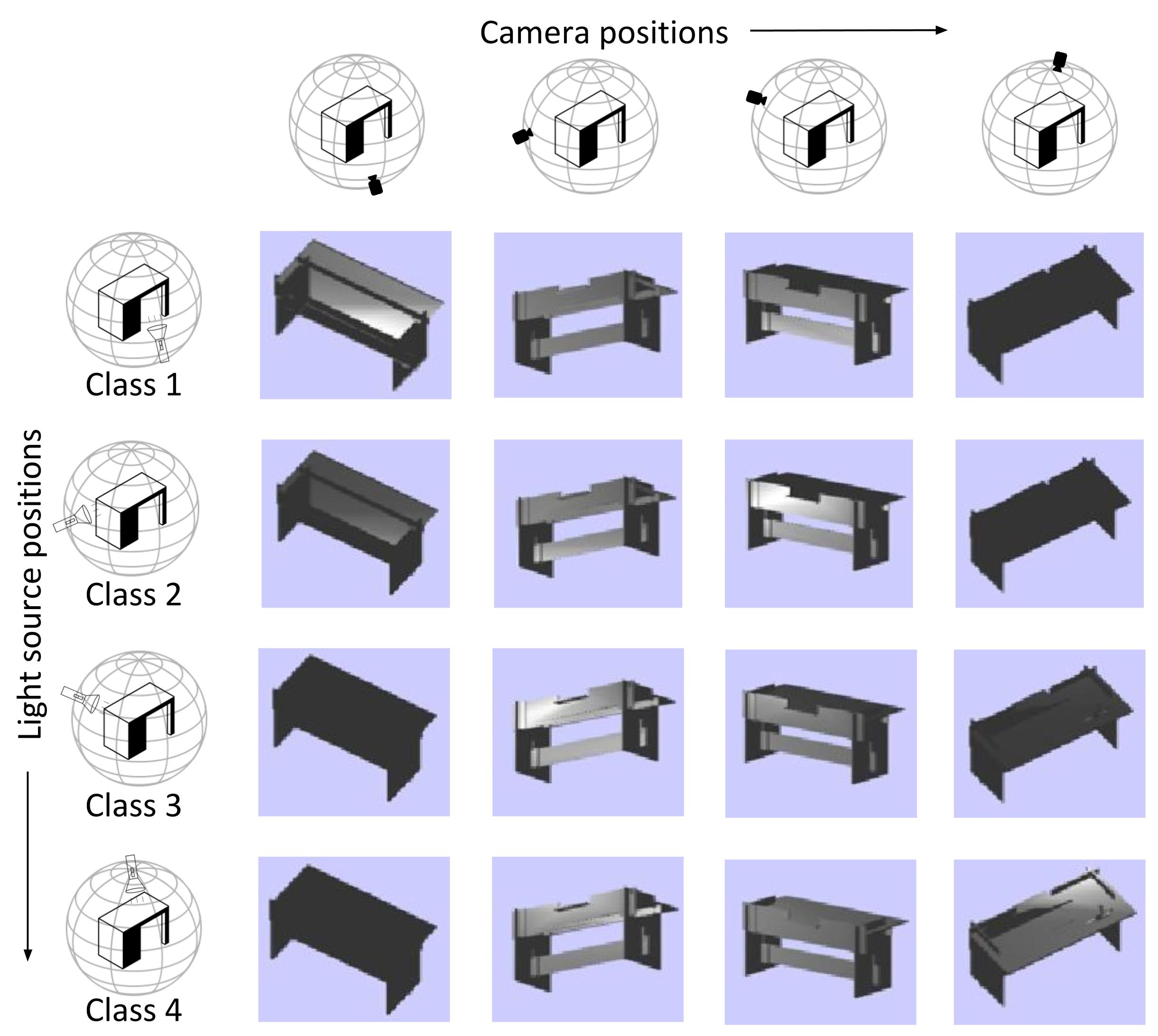}
    \caption{Light source localization example: The different classes are shown in the first column, identified by the elevation of the light source position. The azimuth is varied randomly within each class. The camera positions from which views have been rendered are shown in the top row. Each rendering of the table, indexed by the corresponding class and camera positions, is shown. For class 1, the light source is placed below the object. This is clearly seen in the renderings where images captured from below or towards the sides of the object are lighted and the top view is completely dark. On the other extreme, for class 4, the light source is placed above the table. This is also witnessed in the renderings of the table where the top view is maximally lighted and the bottom view is completely dark.}
    \label{fig:light-source-loc-example}
\end{figure}

\section*{Faster convergence of Sidekick Policy Learning}
As discussed in the main paper, sidekicks lead to faster convergence and better policies by exploiting full observability at training time and guiding the main agent's training. In Figure~S3, we demonstrate the faster convergence in validation errors observed during 1000 epochs of training.
\begin{figure}
    \centering
    \includegraphics[width=\textwidth]{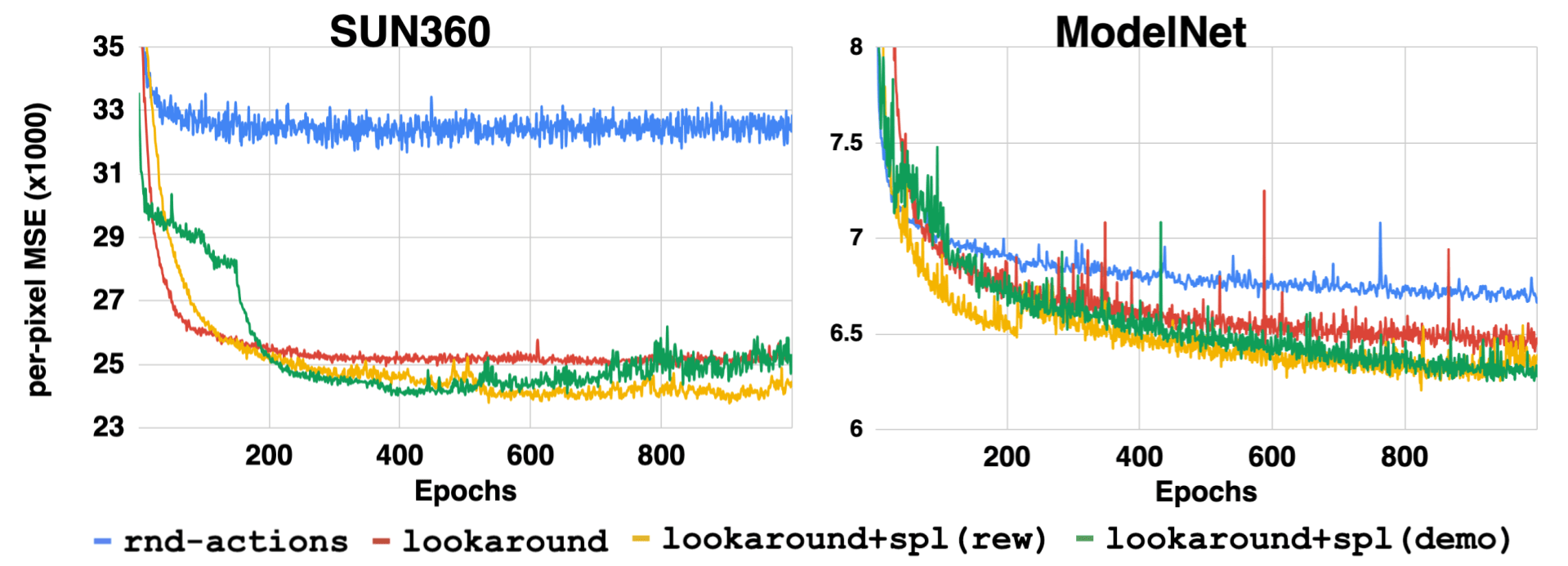}
    \caption{Convergence of sidekick policy learning:  sidekicks lead to faster convergence and better performance. Results are averaged over three different runs.}
    \label{fig:sidekicks-validation}
\end{figure}

\section*{Active observation completion for longer target budgets}
\begin{figure}
    \centering
    \includegraphics[width=\textwidth]{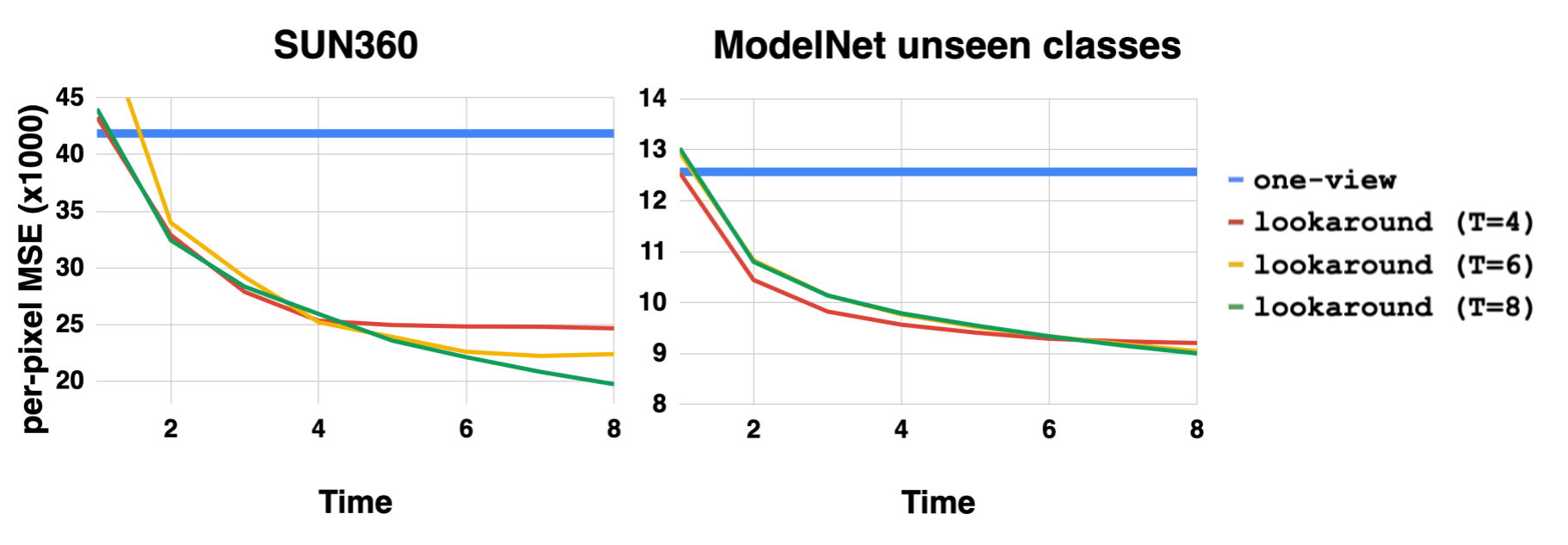}
    \caption{Training on different target budgets $T$: Agents trained on longer episodes consistently improve performance over the entire duration of the episode. Agents trained for shorter time-horizons (\texttt{\small lookaround (T=4, 6)}) naturally tend to saturate in performance earlier, since they target a shorter budget. Agents trained on longer time-horizons \texttt{\small lookaround (T=8)} converge more slowly initially, but eventually outperform the other agents as they approach their target budget.}
    \label{fig:ltla-longer}
\end{figure}

Figure~\ref{fig:ltla-longer} shows that training for longer glimpse budgets does indeed lead to better performance over time.  We train three \texttt{\small look-around} models for budgets $T=4, 6, 8$. As we can see, models trained for longer episodes consistently lead to better performance over time, which is a natural outcome since each agent trains for the budget it is given.

\section*{Additional examples of observation completion episodes}
We show two more sample episodes of observation completion in Figure~S5.
\begin{figure}
    \centering
    \includegraphics[width=\textwidth, trim={0 0 0 0}, clip]{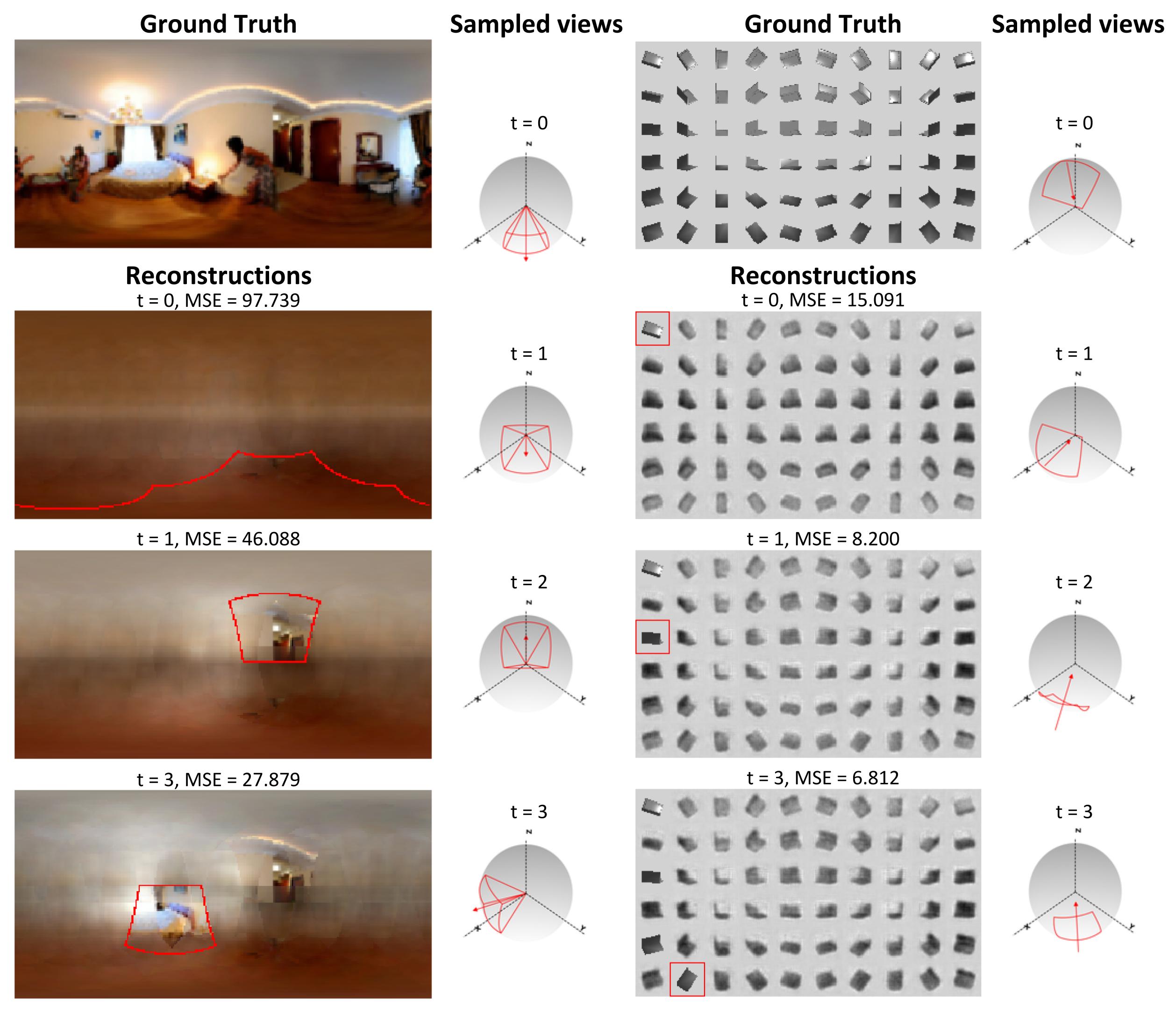}
    \caption{Episodes of active observation completion (continued from Figure~4 in the main paper.)}
    \label{fig:look_around_qual_time_2}
\end{figure}

\section*{GAN refinement for viewgrid visualization}
The goal of our work is to learn useful exploration policies, not to generate photorealistic reconstructions.  Nonetheless, it is valuable to provide good visualizations of the agent's internal state for interpretation of what is being learned.  Therefore, we use recent advances in Generative Adversarial Networks~\cite{goodfellow2014generative} to refine the decoded reconstructions at $T$, both for our method and the baseline. To compute cleaner looking reconstructions, we train the agents for $T=6$. After the final time step, the agent's reconstruction is refined by a pix2pix~\cite{isola2017image} network that is trained to map from the agent's reconstructions to the ground truth viewgrids (see Figure~S6 for examples). 

\section*{Additional loss functions for policy learning}
As mentioned in the policy learning formulation section in Materials and Methods,  the \textsc{act} term additionally includes a loss to update the learned value network (weights $W_{c}$).  
\begin{equation}
\label{eqn:critic_update}
\Delta W_{\{s, f, r, c\}} = -\nabla_{\{s, f, r, c\}} \frac{1}{n}\sum_{i=1}^{n}\sum_{t=1}^{T-1}\bigg(v_{t}^{i} - \sum_{t^{'}=t}^{T-1} r_{t^{'}}^{i}\bigg)^{2},
\end{equation}
where ${n}$ is the number of data samples and $v_{t}^{i} = b(s_{t}^{i})$ is the value estimated by the value network at time $t$ for the $i^{th}$ data sample. 
We additionally include a standard entropy term to promote diversity in action selection and avoid converging too quickly to a suboptimal policy. The loss term and the corresponding weight update (on $W_{a}, W_{r}, W_{f}, W_{s}$) are as follows:
\begin{equation}
\label{eqn:entropy_term}
\begin{split}
L_{ent} = \frac{1}{n}\sum_{i=1}^{n}\sum_{t=1}^{T-1}\bigg(\sum_{a \epsilon \mathcal{A}}\pi(a | s_{t}^{i})\text{log}~\pi(a | s_{t}^{i})\bigg) \\
\Delta W_{\{a, r, f, s\}} = -\nabla_{\{a, r, f, s\}} L_{ent}.
\end{split}
\end{equation}

\begin{figure}
    \centering
    \includegraphics[width=\textwidth, trim={0 0 0 0}, clip]{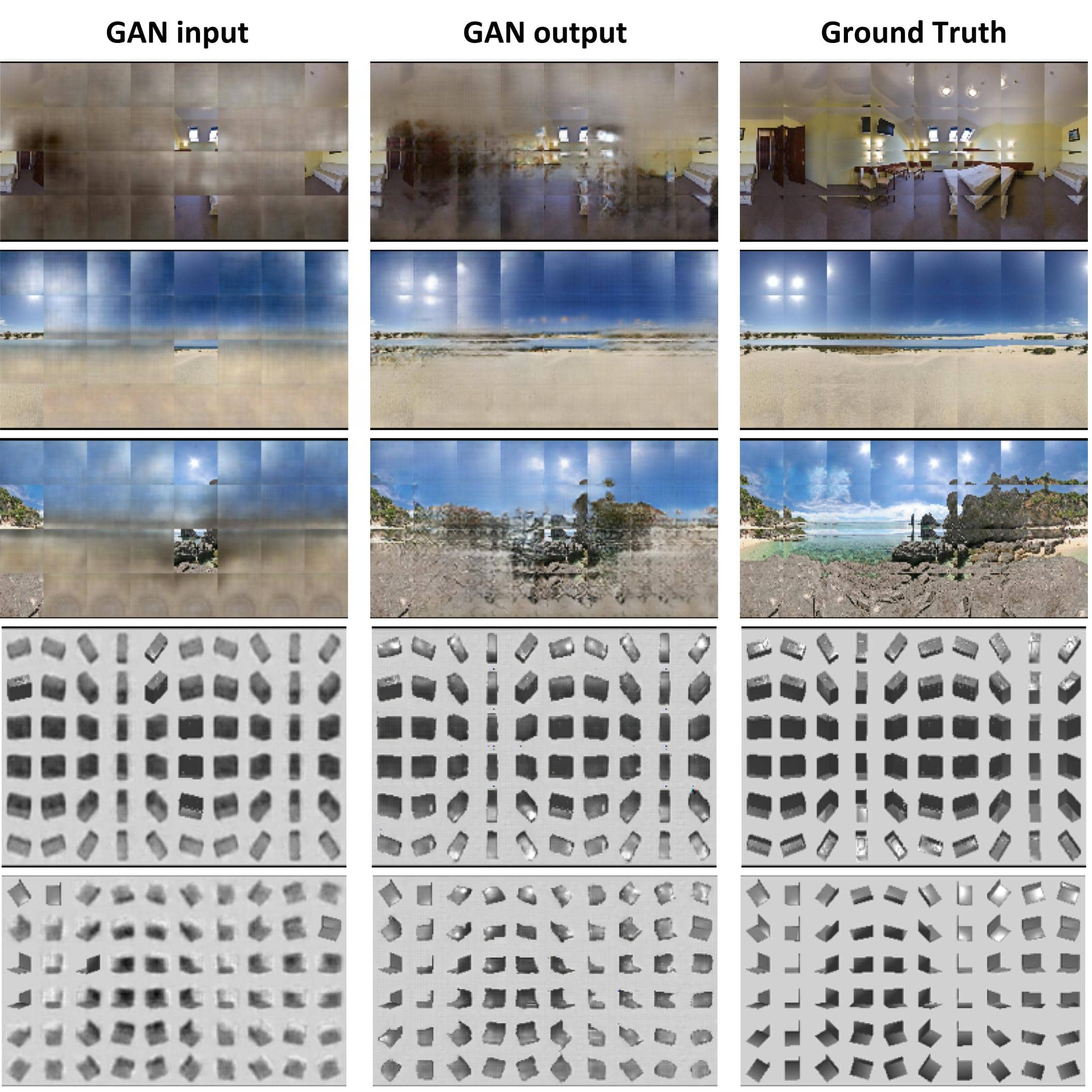}
    \caption{GAN refinement: Given reconstructed viewgrids (1st column) as domain A and ground truth viewgrids (3rd column) as domain B, we train a pix2pix~\cite{isola2017image} network that maps from domain A to domain B. The GAN predictions are shown in column 2. As we can see, the GAN is able to use the high level semantic structure learned by the agent and generate high quality textures using prior knowledge.}
    \label{fig:gan_refinement}
\end{figure}
\end{document}